%% file: main.tex
\documentclass[10pt,twocolumn,letterpaper]{article}

\usepackage[pagenumbers]{cvpr} % To force page numbers, e.g. for an arXiv version
\usepackage{makecell} % Add this package
\usepackage{xcolor}  

\usepackage{bbm}
\definecolor{cvprblue}{rgb}{0.21,0.49,0.74}
\usepackage[pagebackref,breaklinks,colorlinks,allcolors=cvprblue]{hyperref}
\usepackage{graphicx} 
\def\paperID{***} % *** Enter the Paper ID here
\def\confName{CVPR}
\def\confYear{2025}

\title{Enhanced Multi-View Pedestrian Detection Using Probabilistic Occupancy Volume}

\author{Reef Alturki\hspace{1cm}Adrian Hilton\hspace{1cm}Jean-Yves Guillemaut\\
Centre for Vision, Speech and Signal Processing, University of Surrey.\\
{\tt\small \{r.alturki, a.hilton, j.guillemaut\}@surrey.ac.uk}
}
\begin{document}
\maketitle
\input{sec/0_abstract}

\input{sec/1_intro}
\input{sec/relatedwork}
\input{sec/method}

\input{sec/results}

{
    \small
    \bibliographystyle{ieeenat_fullname}
    \bibliography{main}
}


\end{document}

%% file: sec/0_abstract.tex
\begin{abstract}

Occlusion poses a significant challenge in pedestrian detection from a single view. To address this, multi-view detection systems have been utilized to aggregate information from multiple perspectives. Recent advances in multi-view detection utilized an early-fusion strategy that strategically projects the features onto the ground plane, where detection analysis is performed. A promising approach in this context is the use of 3D feature-pulling technique, which constructs a 3D feature volume of the scene by sampling the corresponding 2D features for each voxel. However, it creates a 3D feature volume of the whole scene without considering the potential locations of pedestrians. In this paper, we introduce a novel model that efficiently leverages traditional 3D reconstruction techniques to enhance deep multi-view pedestrian detection. This is accomplished by complementing the 3D feature volume with probabilistic occupancy volume, which is constructed using the visual hull technique. The probabilistic occupancy volume focuses the model's attention on regions occupied by pedestrians and improves detection accuracy. Our model outperforms state-of-the-art models on the MultiviewX dataset, with an MODA of 97.3\%, while achieving competitive performance on the Wildtrack dataset.
\end{abstract}

%% file: sec/1_intro.tex
\section{Introduction}
\label{sec:intro}

Over the past few years, object detection has gained significant interest in the research community, as it plays an essential role in many high-level computer vision tasks. Extensive research has been conducted on object detection using single cameras. Nevertheless, object detection remains challenging in crowded, highly cluttered, and large-scale environments due to frequent occlusions that prevent objects from being fully captured. As a result, detection performance may suffer from increased missed detections.

To overcome occlusion, multi-camera setups with overlapping fields of view are often used, providing comprehensive coverage from multiple perspectives that helps mitigate occlusions, as objects occluded in one camera's view can be captured by another \cite{ref37}. Camera calibrations are typically provided to enable the aggregation of data from multiple perspectives. 

\begin{figure}[t]
    \centering
    \includegraphics[width=0.47\textwidth]{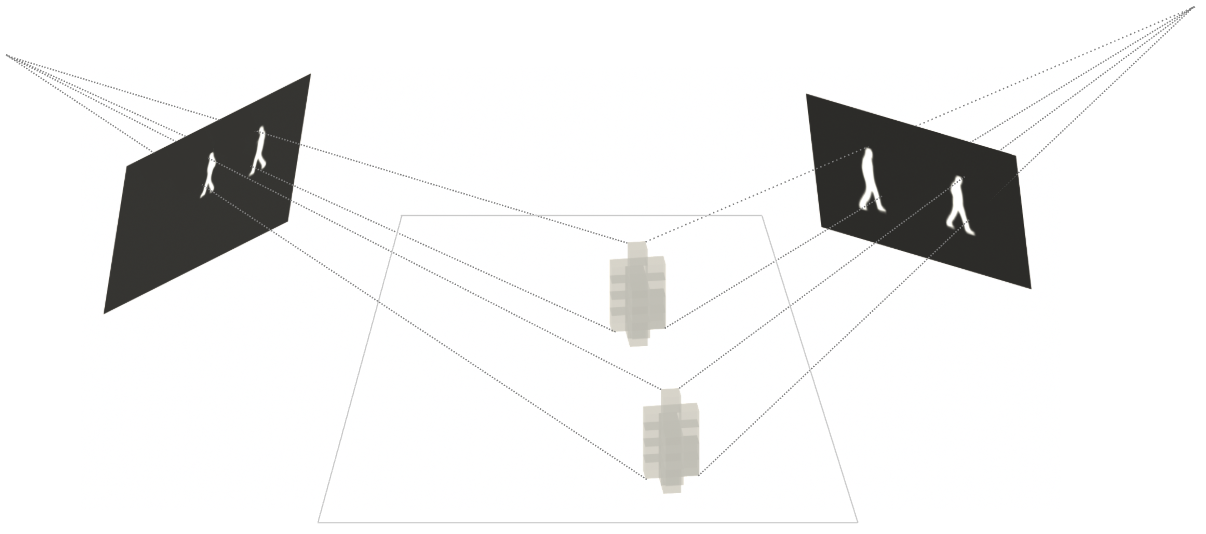} 
    \caption{An example illustrating the occupancy volume of the scene, reconstructed using the visual hull technique, which highlights the voxels corresponding to the regions with high probability of being occupied by pedestrians.}
    \label{fig:fig}
\end{figure}

Recent approaches in multi-view pedestrian detection project the feature maps extracted from all views into the Bird's Eye View (BEV), and then conduct the detection \cite{ref28, ref29}. However, these approaches are based on perspective projection, which results in severe distortion on the ground plane. This distortion causes the pedestrian's features to spread out from their actual locations resembling shadows, thereby impacting the detection of pedestrians located farther away. Additionally, it can lead to the loss of information along the pedestrian's body \cite{ref70}. 

To overcome these limitations, \cite{ref39, ref70} used the 3D feature-pulling method \cite{ref40}, initially proposed as a lifting technique for autonomous vehicle perception, to project the features to a unified 3D space. This method creates a unified 3D feature volume representing the scene by sampling the features from the images for each voxel, effectively addressing the issues arising from perspective projection. When this method is applied to images captured from overlapping camera views, it performs triangulation of image features at the voxel granularity \cite{ref39}. However, while it produces a 3D feature volume of the entire scene, it does not specifically account for the potential locations of pedestrians. This results in insufficient focus on the potential locations occupied by pedestrian, which can hinder its effectiveness for pedestrian detection.

Shape-from-silhouettes technique is a specific category of object reconstruction that estimates the shape of an object by intersecting the volume regions defined by silhouettes from multiple views. Each silhouette confines the object to a specific region in 3D space forming cones. The intersection of cones estimates the volumetric description of the object. The Visual Hull (VH) of an object is defined as the largest possible volume that represents the given silhouettes of an object by intersecting these cone-shaped regions. It provides the maximum boundary of the object's shape, which becomes increasingly accurate with more silhouette images are used \cite{ref67}. 

In this paper, we propose a new model that efficiently incorporates traditional 3D reconstruction techniques with deep learning to enhance multi-view pedestrian detection. Specifically, leveraging the 3D feature-pulling method, the model creates a unified 3D feature volume representing the scene using feature maps from individual views, while incorporating a probabilistic occupancy volume constructed using the visual hull technique \cite{ref67}, which provides complementary information about the locations of pedestrians in the scene. An example of the visual hull of pedestrians is presented in Figure \ref{fig:fig}. To reconstruct the visual hull of pedestrians, we extract high-quality masks from individual views and employ the 3D feature-pulling method to these mask images to construct the visual hull. The visual hull provides the model with a more precise estimate of voxel occupancy by pedestrians, refining its focus on regions occupied by pedestrians, thereby limiting the area of interest to relevant spatial locations. The visual hull is multiplied by the 3D feature volume to give more weight to voxels occupied by pedestrians, the result is concatenated back with the 3D feature volume. Furthermore, it is important to note that the visual hull technique is independent of learnable parameters. As a result, it does not introduce significant overhead in terms of computational complexity, making it an attractive solution for real-time applications.

The main contributions of this paper are as follows:

\begin{itemize}
    \item We propose a  model that leverages a traditional 3D reconstruction technique to enhance deep multi-view pedestrian detection, a novel technique that has not been explored in existing literature.

    \item We propose an efficient method for aggregating the reconstructed pedestrians with the unified scene representation, focusing the model on the potential locations of pedestrians while maintaining stable feature representation.
    
    \item We evaluate the effectiveness of our model both quantitatively and qualitatively by comparing it to state-of-the-art methods. On MultiviewX dataset, our model outperforms existing methods across all metrics, achieving an MODA of 97.3\%. On Wildtrack dataset, our model stands among the top performers and demonstrate competitive performance.

\end{itemize}

%% file: sec/relatedwork.tex
\section{Related Work}
\label{sec:related work}

\textbf{Single-view detection}
The field of object detection has undergone considerable evolution over the years. With the rise of deep learning, object detection methods are typically categorized into two primary types: two-stage and one-stage detectors. Two-stage detectors, such as Fast R-CNN\cite{ref44} and Faster R-CNN \cite{ref21}, first extract regions of interest (RoIs), which are then passed into a network for classification and regression. Mask-RCNN \cite{ref13} builds upon Faster R-CNN by incorporating a mask prediction head, enabling it to simultaneously detect objects and predict their masks. On the other hand, one-stage detectors \cite{ref46, ref15, ref47, ref17} skip the region of interest extraction step, directly classifying and regressing the candidate anchor boxes. While general object detection focuses on detecting various objects, several methods were specifically developed for pedestrian detection \cite{ref81, ref82}. Moreover, occlusion remains a critical challenge in pedestrian detection. To tackle this challenge, part-based models have been developed to address the partial occlusion problem \cite{ref84, ref85}. However, the challenge of occlusion remains a significant obstacle in pedestrian detection in a single-view setup, particularly in crowded scenes where pedestrians frequently overlap.

\begin{figure*}[t]
    \centering
    \includegraphics[width=0.99\textwidth]{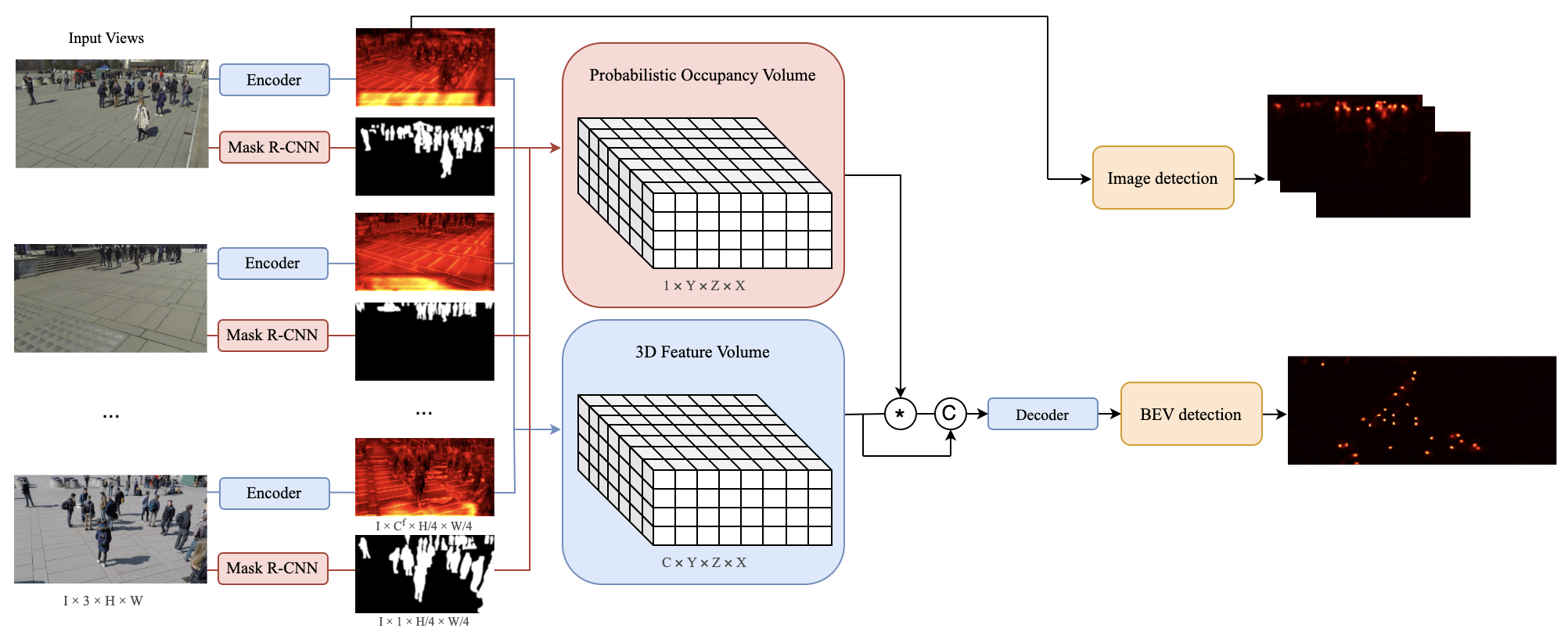} 
    \caption{Overview of our model pipeline. The input views are fed into an encoder to extract feature maps, and Mask R-CNN is applied to the input views to yield silhouettes for pedestrians. The 3D feature-pulling is applied to the feature maps to create the 3D feature volume, it is also applied to Mask R-CNN to compute the probabilistic occupancy volume, which is multiplied with the 3D feature volume and the result is concatenated to the 3D feature volume. The resulting feature is compressed in the vertical dimension and fed into the decoder. $C$ denotes concatenation while * represents element-wise multiplication.}
    \label{fig:fig4}
\end{figure*}

\medskip

\noindent\textbf{Multi-view detection} Multi-view pedestrian detection systems typically use synchronized frames from multiple calibrated cameras that cover overlapping regions, which are then aggregated to perform pedestrian detection. Early approaches \cite{ref35, ref36} were focused on the probabilistic modeling of objects before the advancements brough by deep learning. Several methods \cite{ref33, ref34} used conditional random field (CRF) for aggregating information from multiple views. In \cite{ref83}, the visual hull technique was employed for counting people in crowds. With the advent of deep learning, a new generation of methods emerged. MVDet \cite{ref28} employs a fully convolutional approach, extracting feature maps from the input images and projecting them onto a common ground plane using perspective transformation,  and then applying convolutions with a large receptive field to effectively capture neighboring locations from multiple camera views. This method led to significant improvements and established itself as the foundation for all subsequent approaches. However, its main limitation is the distortions introduced by the perspective transformation, causing pedestrian features to spread out from their locations on the ground plane, resembling shadows of the actual objects \cite{ref29}. Various approaches have been proposed to address this limitation \cite{ref29, ref31, ref30}. MVDeTr \cite{ref29} proposed a deformable attention mechanism to enable adaptive aggregation of features from different cameras and positions, allowing shadow-like features to be relocated to their original locations. Moreover, they introduced view-consistent data augmentation techniques, including cropping, flipping, and scaling to prevent overfitting while maintaining multi-view consistency. SHOT \cite{ref31} proposed projecting the feature maps onto a stack of homographies at different heights to approximate the 3D projection, with a soft selection module predicting the likelihood score of each homography transformation for every pixel according to its semantics. MVTT \cite{ref30} utilizes single-view detection results to encode the full features of each pedestrian by applying Region of Interest (RoI) pooling to the detected bounding boxes, followed by projecting the encoded features onto the predicted foot positions on the ground plane. On the other hand, 3DROM \cite{ref32} proposed a data augmentation technique that involves adding additional occlusions by randomly placing 3D cylindrical objects on the ground plane, thereby reducing overfitting and making the model more robust to occlusion. MVFP \cite{ref70} employs the 3D feature-pulling strategy \cite{ref40} which constructs a 3D feature volume of the scene by sampling the respective 2D features for each valid voxel. This approach effectively mitigates feature loss along the pedestrian’s body that occurs in methods relying on perspective projection.

Although MVFP achieves state-of-the-art results, it creates a unified 3D feature volume of the whole scene without considering the potential locations where pedestrians may be located. Moreover, none of the existing methods have integrated the visual hull into a deep learning framework for multi-view detection. Therefore, we introduce a novel approach that integrates traditional 3D reconstruction techniques with deep learning for multi-view pedestrian detection. By using the 3D feature-pulling technique for multi-view aggregation and visual hull reconstruction, our model improves pedestrian localization and boosts detection accuracy.

%% file: sec/method.tex
\section{Methodology}

The overall structure of the proposed model is shown in Figure \ref{fig:fig4}. Given a set of synchronized images from $I$ calibrated cameras, these images are augmented and fed into the encoder to extract the feature maps. Next, the feature maps are projected to a common BEV space using the 3D feature-pulling method \cite{ref40}. This creates a 3D feature volume for each view by sampling the 2D image features for each voxel, then the volumes are aggregated across the views using valid-weighted average, ultimately resulting in a 3D feature volume representing the scene. Moreover, Mask R-CNN \cite{ref13} is applied to the input images to extract high quality masks of the pedestrians, which are used to construct a probabilistic occupancy volume of the pedestrians by employing the 3D feature-pulling method. The probabilistic occupancy volume is computed using the Probabilistic Visual Hull (PVH) method, which provides the probability of occupancy for each voxel and is multiplied with the 3D feature volume to emphasize the voxels occupied by pedestrians. The resulting feature is then concatenated with the original 3D feature volume to account for inaccuracies in camera calibration and ensure a more stable feature representation. The BEV feature is then compressed in vertical dimension and fed through the decoder network.

\subsection{Encoder}
The encoder, based on \cite{ref37}, takes input images from $I$ cameras, each having an input size of  $3 \times H \times W$, and extracts the feature maps using ResNet. The ResNet consists of three blocks that progressively downsample the images by a factor of 2. The output features of each layer are upsampled and concatenated with the output features from the previous layer. These combined features are then passed through a convolutional layer, producing the final feature map with dimensions of $C^{f} \times H^{f} \times W^{f}$, where $H^{f} = H/4$, $W^{f} = W/4$, and $C^{f}$ represents the output feature channels.

\subsection{3D Feature-Pulling}
The features maps from all $I$ cameras are projected to a common BEV space using the 3D feature-pulling method \cite{ref40}, which produces a 3D feature volume representing the scene. The process involves creating a 3D voxel coordinate volume $V_i(x, y, z) \in \mathbb{R}^{Y \times Z \times X}$ for each camera $i$. The dimensions $Y$ and $X$ are determined by the size of the ground plane $[H^g, W^g]$, while $Z$ represents the presumed vertical dimension. The coordinate volume is projected onto each camera’s view using the pinhole camera model \cite{ref55}. The translation between 2D image pixel coordinates $(u, v)$ and 3D
locations $(x, y, z)$ is computed with: 
\begin{equation}
\scalebox{0.9}{$
d \begin{pmatrix} u \\ v \\ 1 \end{pmatrix} = K [ R | t ] \begin{pmatrix} x \\ y \\ z \\ 1 \end{pmatrix} =  P^{(i)} \begin{pmatrix} x \\ y \\ z \\ 1 \end{pmatrix} $}
\end{equation}
where $d$ denotes a scaling factor accounting for the downsampling of the feature map, and $P^{(i)} = K [R | t]$ is a $3 \times 4$ transformation matrix for camera $i$. In particular, $K$ is the intrinsic parameter matrix, while $[R | t]$ denotes the extrinsic parameter matrix, where $R$ specifies the rotation and $t$ specifies the translation. 

To address the issue of 3D voxels lying outside the camera’s frustum, a validity mask $B_{i}$ is
calculated for each voxel with respect to camera $i$. The mask determines if the voxel's projected 2D coordinates fall within the image bounds:
\begin{equation}
\scalebox{0.9}{$
B_{i}(x, y, z) = 
\begin{cases} 
1, & \text{if } u \in [0, W^f] \text{ and } v \in [0, H^f] \\ 
0, & \text{otherwise}
\end{cases}$}
\end{equation}

The validity mask is used to eliminate the features of voxels falling outside the camera's frustum. For each valid voxel, bilinear sampling is used to extract the 2D features $C^f$ sub-pixel accurate to the voxel, resulting in a 3D feature volume $\mathcal{F}_{i}^{f} \in \mathbb{R}^{C \times Y \times Z \times X}$ for each view $i$:
\begin{equation}
\scalebox{0.9}{$
    \mathcal{F}_{i}^{f} = \begin{cases} 
    \mathrm{sampling}(C_i^{f}(u,v)) & \text{if } B_{i}(x, y, z) = 1 \\
    0 & \text{otherwise}
    \end{cases}$}
\end{equation}

To aggregate the 3D feature volumes across the views, we apply average pooling on valid voxels across the different views. This pooling operation results in a single 3D feature volume $\mathcal{F}^{f} \in \mathbb{R}^{C \times Y \times Z \times X}$, which serves as a unified representation of the scene.

\subsection{Probabilistic Occupancy Volume}
We leverage the visual hull technique \cite{ref67} to assist the model in identifying the locations of
pedestrians in the scene, which is represented in the form of a probabilistic occupancy volume. To achieve this, we use Mask R-CNN \cite{ref13} to obtain high-quality
masks of pedestrians, which serve as silhouettes for visual hull reconstruction. To ensure proper alignment with the feature maps, the silhouette images are downsampled to match their size, with blurring applied first to reduce aliasing artifacts in downsampling. The 3D feature-pulling method is then applied to the silhouette images to compute the visual hull. Specifically, a 3D voxel coordinate volume is created for each view, and each voxel samples the corresponding 2D point from the silhouette images using bilinear sampling, resulting in a 3D occupancy volume $\mathcal{F}_{i}^{o} \in \mathbb{R}^{1 \times Y \times Z \times X}$ for each view $i$. Next, the visual hull is created by evaluating the occupancy of each voxel. A voxel $(x, y, z)$ is considered part of the visual hull if the total number of non-zero values sampled across all valid views is equal to the number of views where the voxel’s projection lies within the camera’s frustum. This condition is illustrated as:
\begin{equation}
\scalebox{0.84}{$
V_{\text{h}}(x, y, z) = 
\begin{cases}
1 & \text{if } \sum_{i=1}^{N} \mathbbm{1}\left(\mathcal{F}_i^{o}(x, y, z) > 0 \right) = N_{\text{v}}(x, y, z), \\
0 & \text{otherwise}
\end{cases}
$}
\end{equation}
where $\mathbbm{1}\left(\mathcal{F}_i^{o}(x, y, z) > 0 \right)$ evaluates to 1 if the sampled value is non-zero. The total number of views where a voxel’s projection lies within the camera’s frustum is calculated using validity mask values across all views. For each voxel $(x, y, z)$, this is given by:
\begin{equation}
\scalebox{0.9}{$
N_{\text{v}}(x, y, z) = \sum_{i=1}^{N} B_i(x, y, z)$}
\end{equation}
Once the voxels that are part of the visual hull are identified, we use the Probabilistic Visual Hull (PVH) \cite{ref71}, a variant of the traditional visual hull. Instead of relying on binary values \{0, 1\}, PVH calculates the occupancy probability for a given voxel by multiplying the sampled values across all valid views, as follows:
\begin{equation}
\scalebox{0.9}{$
\mathcal{F}^{o} = 
\begin{cases}\prod_{i=1}^{N_{\text{v}}(x, y, z)} \mathcal{F}_i^{p}(x, y, z) & \text{if } V_{h}(x, y, z) = 1 \\
0 & \text{otherwise}
\end{cases}
$}
\end{equation}
This results in a 3D volume, $\mathcal{F}^{o} \in \mathbb{R}^{1 \times Y \times Z \times X}$, that provides a refined estimate of voxel occupancy. We multiply the occupancy volume $\mathcal{F}^{o}$ by the 3D feature volume $\mathcal{F}^{f}$, ensuring that the features from the voxels occupied by people are preserved and given more weight. This selective weighting enhances the model's capability to detect and precisely localize people on the ground plane. The result of the multiplication is concatenated to the original 3D feature volume. The concatenation of the occupancy volume with the original 3D feature volume aids in addressing misalignments from inaccurate calibration data, ensuring more stable feature representation. Finally, the vertical dimension $Z$ is compressed using a convolutional layer, producing a 2D BEV feature map, $C^g \times H^g \times W^g$, optimized for localization on the ground plane.

\subsection{Decoder}
Prior approaches \cite{ref37, ref38, ref39} have employed Feature Pyramid Networks on the BEV feature to introduce a large receptive field for each location, enabling effective aggregation of location and appearance features of pedestrians. Our work builds on the approaches \cite{ref38, ref39}, which utilize a ResNet-18 decoder. In this framework, the BEV feature is downsampled by a factor of 2 in each layer of the ResNet. Each layer's output is upsampled to match the size of the previous larger output. Afterward, the two features are concatenated along the channel dimension and passed through a deformable convolution layer \cite{ref69}, which alleviates spatial misalignment across scales in the fusion stage, leading to more accurate detection. The decoded output from the feature pyramid maintains the same dimensions as the input, while having a larger receptive field for each grid location.

\subsection{Heads and losses}
To compute the Probabilistic Occupancy Map (POM) on the ground plane, we adopt the CenterNet \cite{ref80} architecture. In particular, we apply a center prediction head that reduces the BEV to a size of 
 $1 \times H^g \times W^g$, producing a heatmap representing the POM on the ground plane. We also apply an offset prediction head to produce an offset map, $2 \times H^g \times W^g$, to recover the omitted factional part during the downsampling process. Focal Loss \cite{ref59} is used to optimize the center head, while L1 Loss is used to optimize the offset head. Moreover, we incorporate detection heads for the image features that estimate the 2D bounding boxe centers and the approximated foot position at the bottom-center of the bounding box. This enables the model to allocate more attention to areas where pedestrians are located.

%% file: sec/results.tex
\begin{table*}[t]
\centering
    \renewcommand{\arraystretch}{1.1} % Adjust row height as needed
\label{tab:results}
    \begin{tabular}{@{}lcccccccc@{}}
\toprule
  & \multicolumn{4}{c}{WildTrack} & \multicolumn{4}{c}{MultiviewX} \\
\cmidrule(lr){2-5} \cmidrule(lr){6-9}
 & MODA & MODP & Precision & Recall & MODA & MODP & Precision & Recall \\
\midrule
MVDet    \cite{ref28}  & 88.2 & 75.7 & 94.7 & 93.6 & 83.9 & 79.6 & 96.8 & 86.7 \\
SHOT   \cite{ref31}      & 90.2 & 76.5 & 96.1 & 94.0 & 88.3 & 82.0 & 96.6 & 91.5 \\
MVDeTr  \cite{ref29}  & 91.5 & \underline{82.1} & \underline{97.4} & 94.0 & 93.7 & 91.3 & \textbf{99.5} & 94.2 \\
MVAug  \cite{ref75}   & 93.2 & 79.8 & 96.3 & \underline{97.0} & 95.3 & 89.7 & \underline{99.4} & 95.9 \\
3DROM \cite{ref32}  & 93.5 & 75.9 & 97.2 & 96.2 & 95.0 & 84.9 & 99.0 & 96.1 \\
EarlyBird  \cite{ref37}    & 91.2 & 81.8 & 94.9 & 96.3 & 94.2 & 90.1 & 98.6 & 95.7 \\
TrackTacular \cite{ref39}  & 92.1 & 76.2 & 97.0 & 95.1 & \underline{96.5} & 75.0 & \underline{99.4} & 97.1  \\
MVTT     \cite{ref30}   & \textbf{94.1} & 81.3 & \textbf{97.6} & 96.5 & 95.0 & \underline{92.8} & \underline{99.4} & 95.6 \\
MVFP    \cite{ref70}  & \textbf{94.1} & 78.8 & 96.4 & \textbf{97.7} & 95.7 & 85.1 & 98.4 & \underline{97.2} \\
\midrule
Ours           &  \underline{93.6} & \textbf{82.4} & 96.6 & \underline{97.0} & \textbf{97.3} & \textbf{95.0} &  \textbf{99.5} & \textbf{97.9} \\
\bottomrule
\end{tabular}
    \caption{Detection performance evaluation on Wildtrack and MultiviewX datasets agaisnt state-of-the-art models. The highest and second-highest performers are denoted by bold and underlined, respectively.} 
    \label{tab:performance_metrics1}
\end{table*}

\begin{table}[t]
    \centering
    \renewcommand{\arraystretch}{1.1} % Adjust row height as needed
    \begin{tabular}{@{}lcccc@{}}
        \toprule
        Configuration & MODA & MODP & Prec. & Recall \\ 
        \midrule
        +VH & \underline{97.1} & \underline{94.8} & \underline{99.4} & \underline{97.7} \\
        + PVH & \textbf{97.3} & \textbf{95.0} & \textbf{99.5} & \textbf{97.9} \\
        \bottomrule
    \end{tabular}
    \captionsetup{justification=raggedright}
    \caption{Comparison of detection performance between the visual hull (VH) and probabilistic visual hull (PVH).}
    \label{tab:ablation_study2}
\end{table}

\section{Experiments}
\subsection{Datasets}

\textbf{Wildtrack Dataset} Wildtrack \cite{ref49} is a dataset captured from a real-world scene using seven static cameras, each recording video at $1080 \times 1920$ resolution and a frame rate of 2 fps. These cameras are synchronized and calibrated with a highly overlapping field-of-view, covering a region of $12 \, \text{m} \times 36 \, \text{m}$. This area is discretized into a $480 \times 1440$ grid, yielding grid cells of $2.5 \, \text{cm} \times 2.5 \, \text{cm}$. Each frame contains an average of 20 pedestrians, and each location is monitored by around 3.74 cameras. The dataset consists of 400 frames, with the first 360 frames used to train the model and the last 40 frames used for testing.

\medskip
\noindent\textbf{MultiviewX Dataset} MultiviewX \cite{ref28} is a synthetic dataset generated using a game engine, designed to have similar properties to Wildtrack. It maintains the same resolution, frame rate, and frame count as Wildtrack, but uses one fewer camera. It covers an area of $16 \, \text{m} \times 25 \, \text{m}$, discretized into a $640 \times 1000$ grid with the same cell size. On average, it captures 40 pedestrians per frame, with around 4.41 cameras covering each location. Similarly, it comprises 400 frames, where the first 360 frames used to train the model, and the subsequent 40 frames used to test the model.

\subsection{Evaluation Metrics}

Multi-view detection systems typically evaluate the POM on the ground plane using Euclidean distance to compare the predicted and ground truth detections \cite{ref49}. We use four evaluation metrics to assess the performance: Multiple Object Detection Accuracy (MODA), Multiple Object Detection Precision (MODP), Precision, and Recall. MODA is computed as:
\begin{equation}
\text{MODA} = 1 - \frac{FP + FN}{N},
\end{equation}
where \( FP \) is the number of false positives, \( FN \) is the number of false negatives, and \( N \) is the number of ground-truth pedestrians. MODP is calculated as:
\begin{equation}
\text{MODP} = \frac{\sum (1 - d[d<t]/t)}{TP},
\end{equation}
where \( d \) represents the distance between a detection and its corresponding ground truth, with $t$ set to $0.5\,\textnormal{m}$ as a threshold to determine true positives, and \( TP \) is the number of true positives. Moreover, Precision is defined as:
\begin{equation}
\text{Precision} = \frac{TP}{TP + FP}, 
\end{equation}
while the Recall is defined as:
\begin{equation}
\text{Recall} = \frac{TP}{N}.
\end{equation}

As done in prior works \cite{ref28, ref29}, we use Multiple Object Detection Accuracy (MODA) as the key performance metric, as it considers both false positives and missed detections.

\subsection{Implementation Details}

The input images have dimensions of $720 \times 1280$ pixels. To prevent overfitting and improve generalization, a series of augmentations are applied during training. Specifically, each image is randomly resized within a scale range of $[0.8,1.2]$, shifted by a random offset from the center, and cropped to preserve its original size \cite{ref29, ref40}. The camera intrinsics matrices are adjusted accordingly to maintain multi-view consistency. In addition, some noise is introduced to the translation vector of the extrinsic matrix to avoid overfitting the detector, following the approach in \cite{ref37, ref39}. For training, we use Adam optimizer with a one-cycle learning rate scheduler, setting the peak learning rate to $1 \times 10^{-3}$. The encoder and decoder, as well as Mask R-CNN, are initialized using weights pre-trained on \textit{ImageNet-1K}.

\subsection{Quantitative Results}

In Table~\ref{tab:performance_metrics1}, the performance of our model is compared to state-of-the-art methods on Wildtrack and MultiviewX datasets. Our model consistently outperforms
existing methods by a large margin across all metrics on MultiviewX dataset, achieving an increase of $+0.8$ in MODA compared to TrackTacular \cite{ref39}. On Wildtrack dataset, our model ranks among the top, securing a strong second place with an MODA of $93.6\%$ and recall of $97.0\%$, in addition to the highest MODP of $82.4\%$. Moreover, the results on MultiviewX dataset show larger improvements across all methods compared to those on Wildtrack dataset. According to \cite{ref39}, the results might imply that the labeling accuracy on Wildtrack could be a restrictive factor, as they annotated using perspective transform \cite{ref49}.

\begin{table}[t]
\centering
\renewcommand{\arraystretch}{1.3} % Adjust row height as needed
\label{tab:results}
\begin{tabular}{@{}p{3.3cm}p{0.8cm}p{0.8cm}p{0.6cm}p{0.8cm}@{}}
    \toprule
    Configuration & MODA & MODP & Prec. & Recall \\
    \midrule
    Basline & 96.8 & \underline{94.7} & \textbf{99.5} & 97.3  \\

    \makecell[l]{+PVH \\\textit{Concat PVH}} & 97.1 &  94.5 &  \underline{99.4} &  \underline{97.7} \\
    \makecell[l]{+PVH \\\textit{Mult PVH}} & 93.8 & 94.3 &  \underline{99.4} & 94.4  \\
    \makecell[l]{+PVH \\\textit{Mult \& Add PVH}}  & \underline{97.2} & 94.4 & \textbf{99.5} &  \underline{97.7} \\
    \makecell[l]{+PVH \\\textit{Mult \& Concat PVH}} & \textbf{97.3} & \textbf{95.0} & \textbf{99.5} & \textbf{97.9}  \\
    \bottomrule
\end{tabular}

\caption{Comparisons of various PVH integration methods (concatenation, multiplication, and addition) with the original features, evaluating their impact on detection performance.}
\label{tab:ablation_study1}
\end{table}

\begin{figure*}[t]
  \centering

  \begin{subfigure}{0.27\textwidth}
    \centering
    \includegraphics[width=\textwidth]{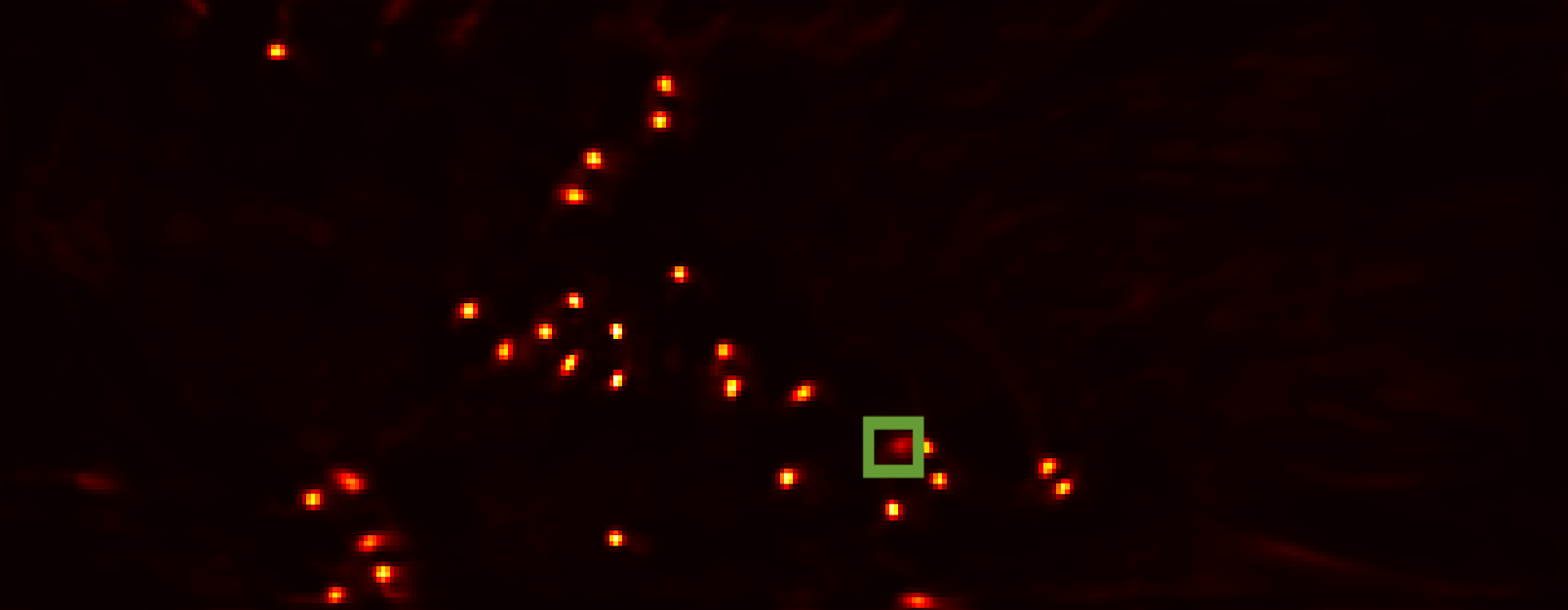}
    \label{fig:prediction_2}
  \end{subfigure} 
  \begin{subfigure}{0.27\textwidth}
    \centering
    \includegraphics[width=\textwidth]{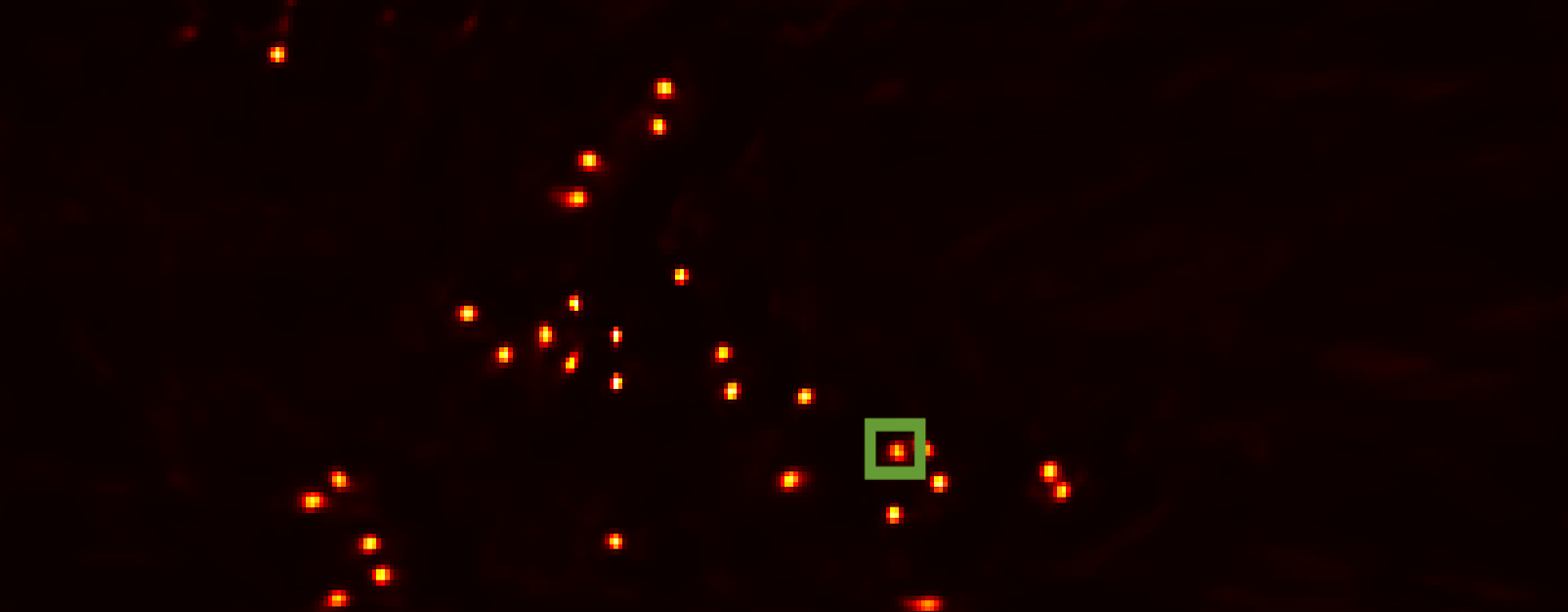}
    \label{fig:ground_truth_2}
  \end{subfigure} 
  \begin{subfigure}{0.27\textwidth}
    \centering
    \includegraphics[width=\textwidth]{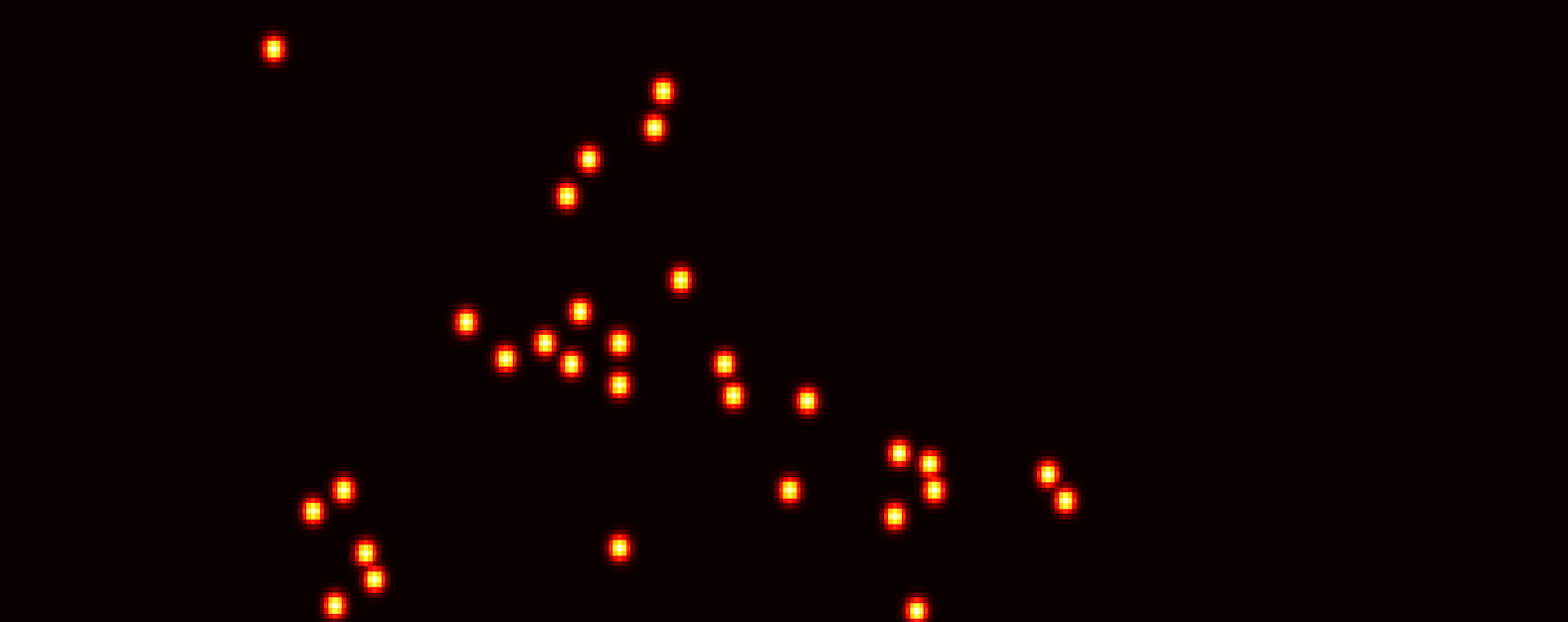}
    \label{fig:third_image_2}
  \end{subfigure}
  
   \begin{subfigure}{0.27\textwidth}
    \centering
    \includegraphics[width=\textwidth]{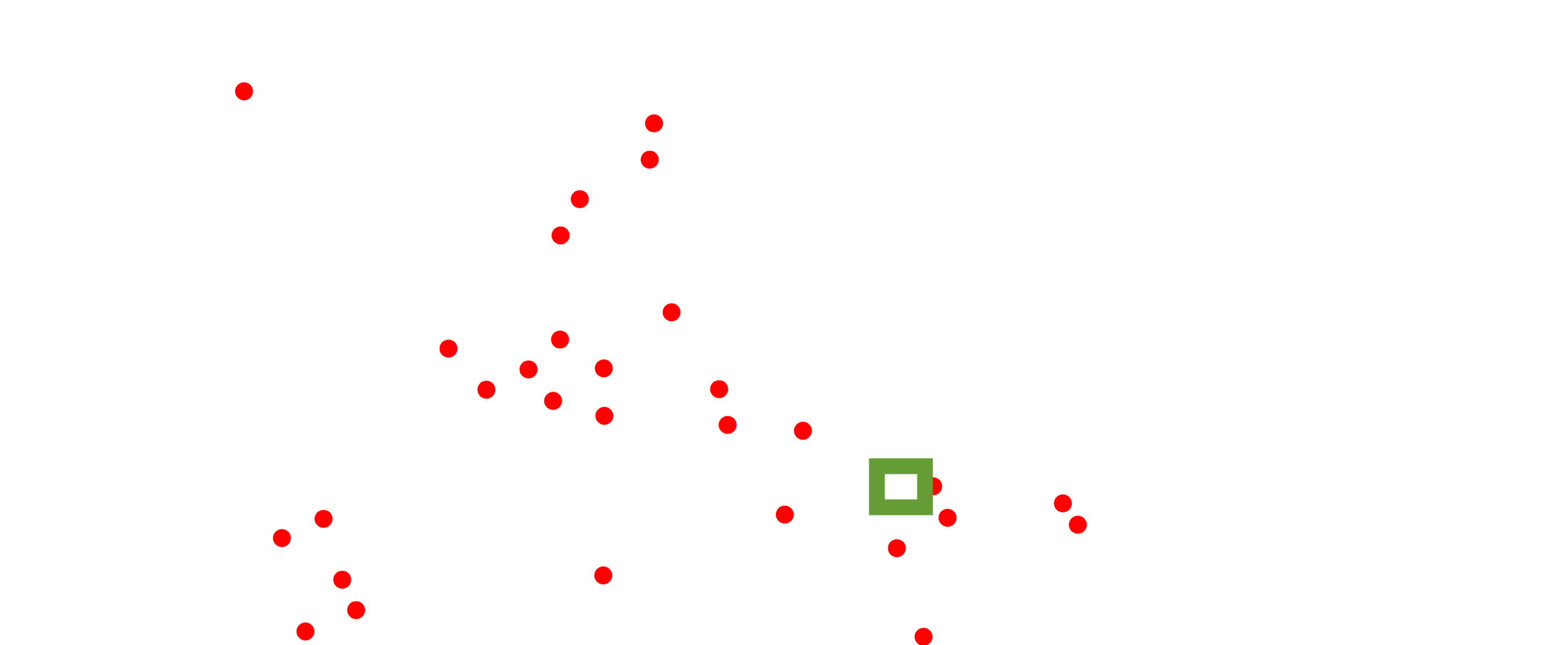}
    \label{fig:prediction_2}
  \end{subfigure} 
  \begin{subfigure}{0.27\textwidth}
    \centering
   \includegraphics[width=\textwidth]{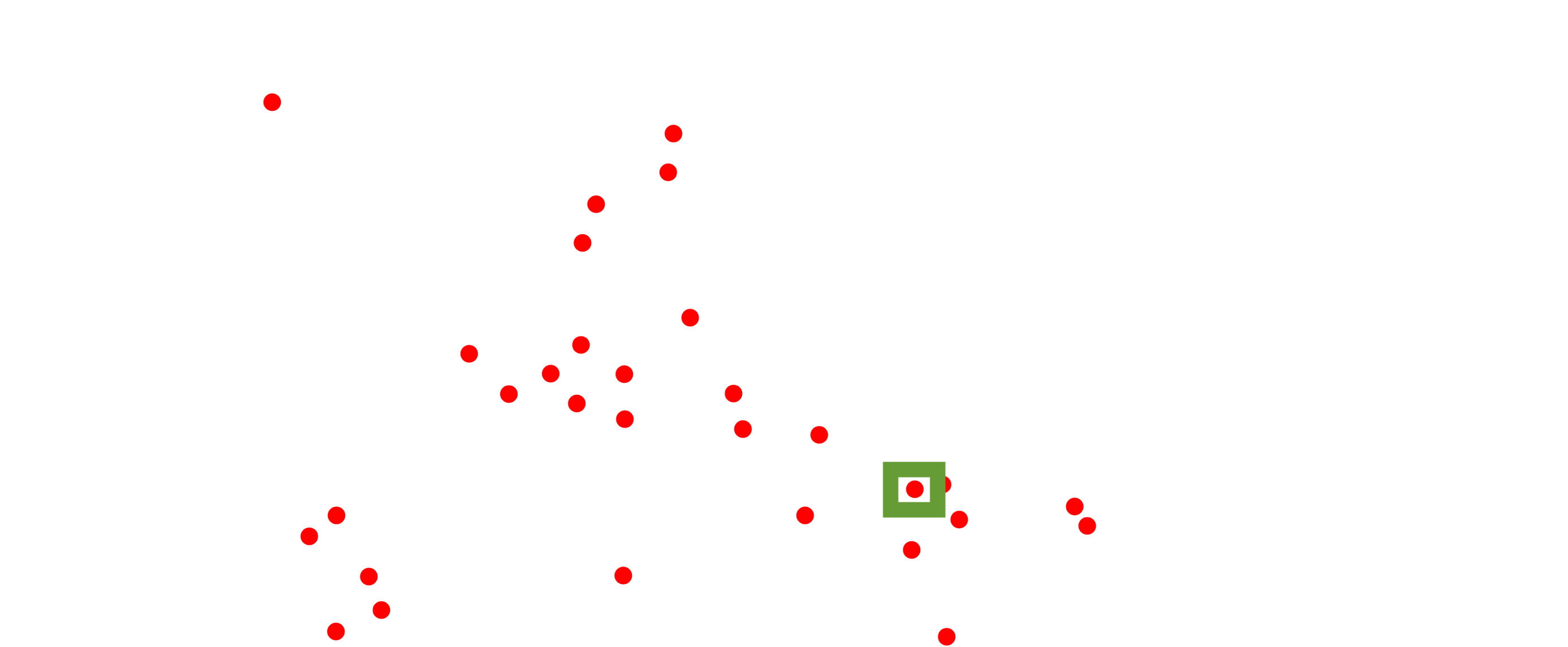}
    \label{fig:ground_truth_2}
  \end{subfigure} 
  \begin{subfigure}{0.27\textwidth}
    \centering
    \includegraphics[width=\textwidth]{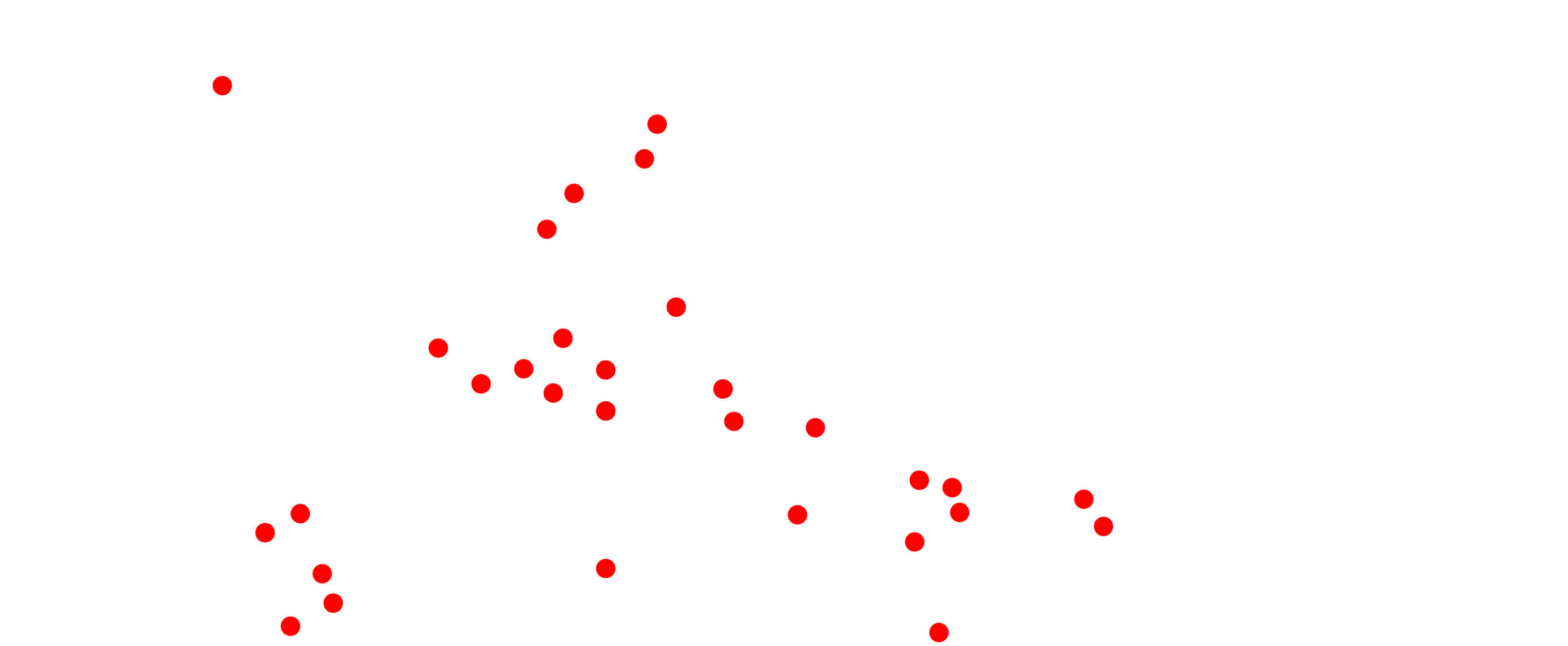}
    \label{fig:third_image_2}
  \end{subfigure}

  \vspace{-2mm} 
  \noindent\textcolor{gray}{\rule[0mm]{14.2cm}{0.1pt}} 
  \vspace{1mm}

    \begin{subfigure}{0.27\textwidth}
    \centering
    \includegraphics[width=\textwidth]{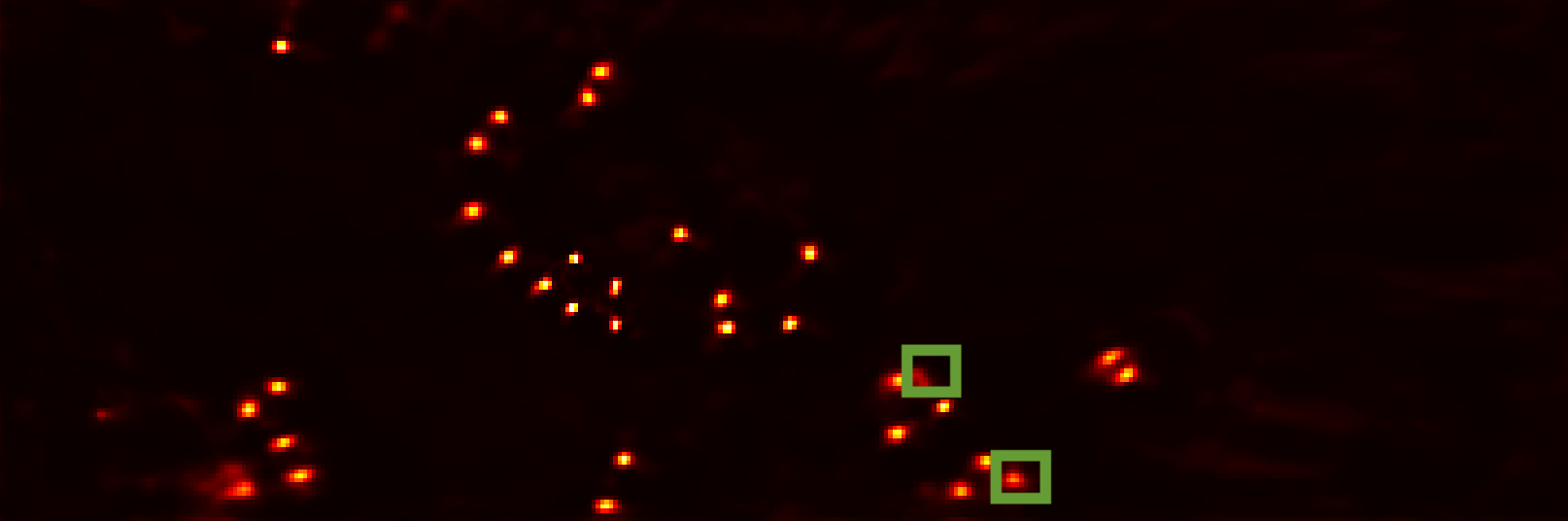}
    \label{fig:prediction_2}
  \end{subfigure} 
  \begin{subfigure}{0.27\textwidth}
    \centering
    \includegraphics[width=\textwidth]{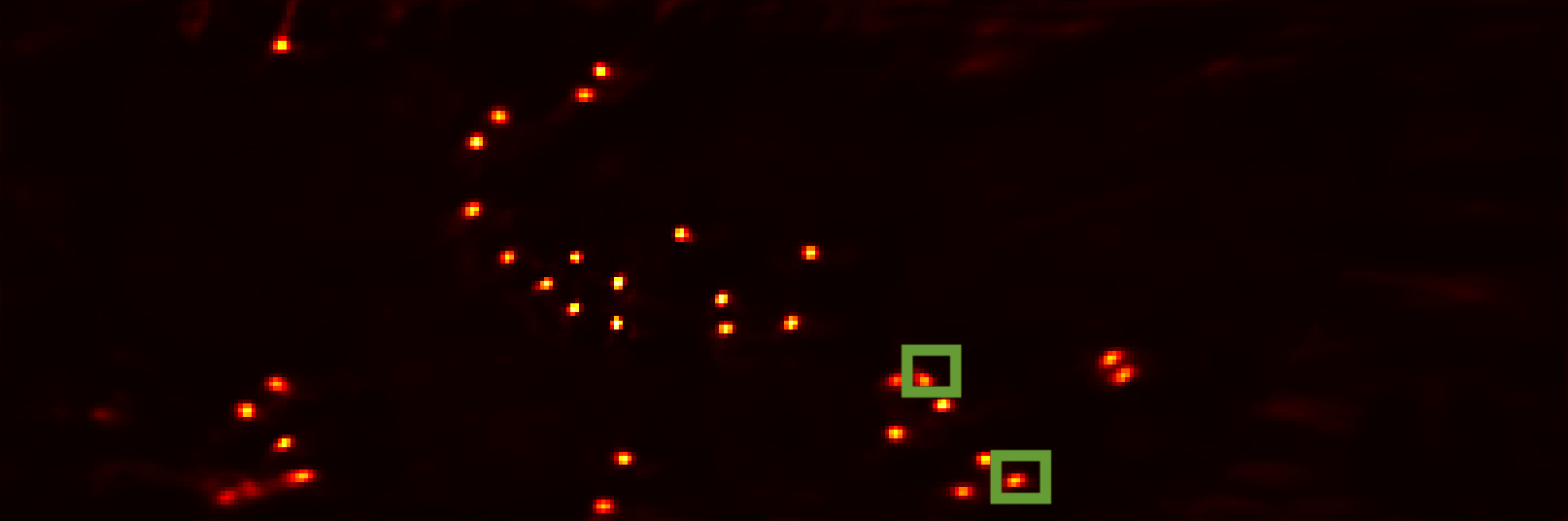}
    \label{fig:ground_truth_2}
  \end{subfigure} 
  \begin{subfigure}{0.27\textwidth}
    \centering
    \includegraphics[width=\textwidth]{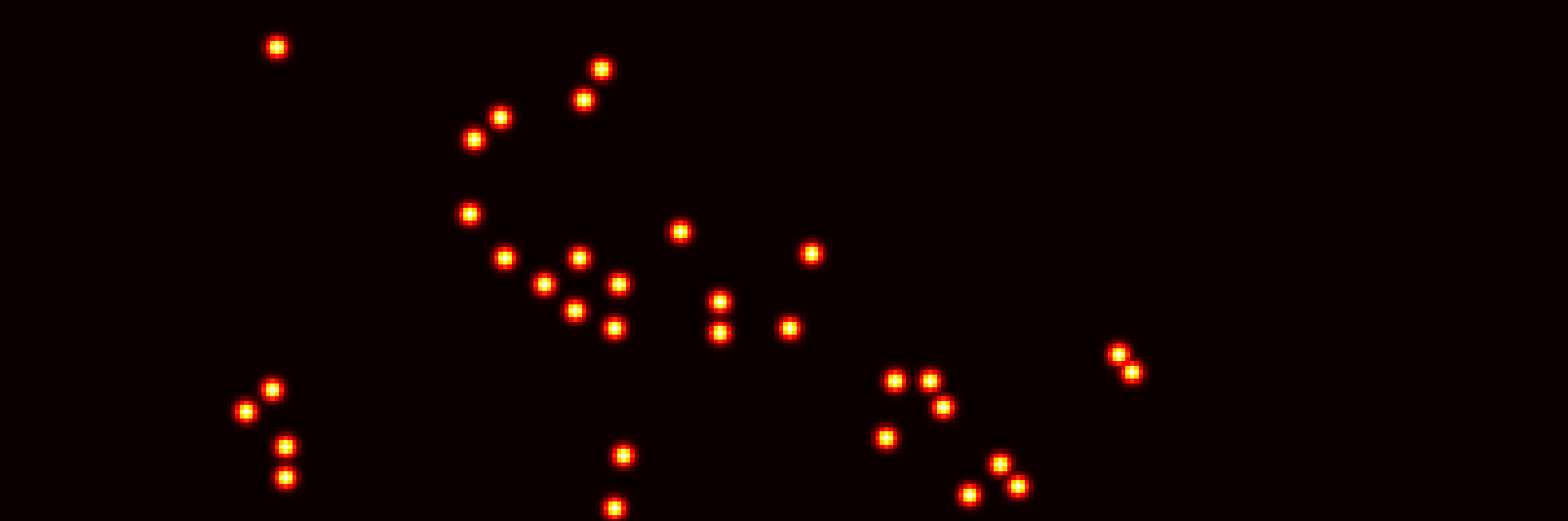}
    \label{fig:third_image_2}
  \end{subfigure}
  
   \begin{subfigure}{0.27\textwidth}
    \centering
    \includegraphics[width=\textwidth]{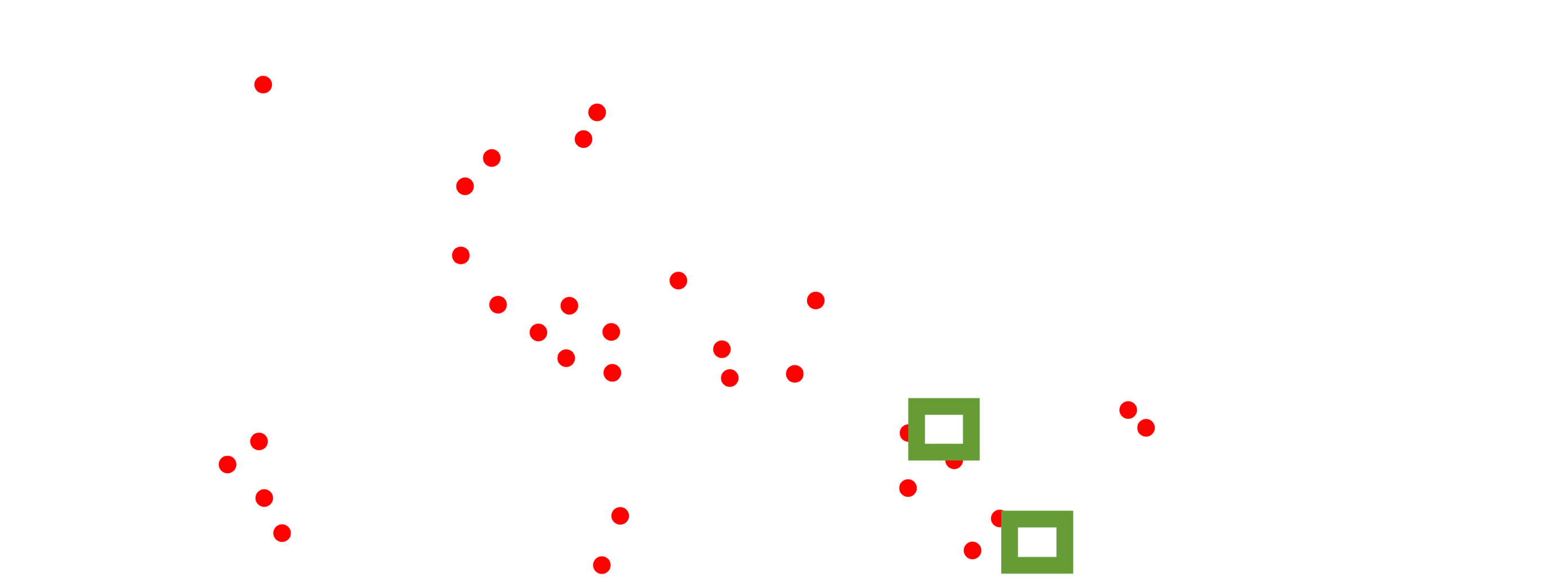}
       \caption{Baseline.}
    \label{fig:prediction_2}
  \end{subfigure} 
  \begin{subfigure}{0.27\textwidth}
    \centering
    \includegraphics[width=\textwidth]{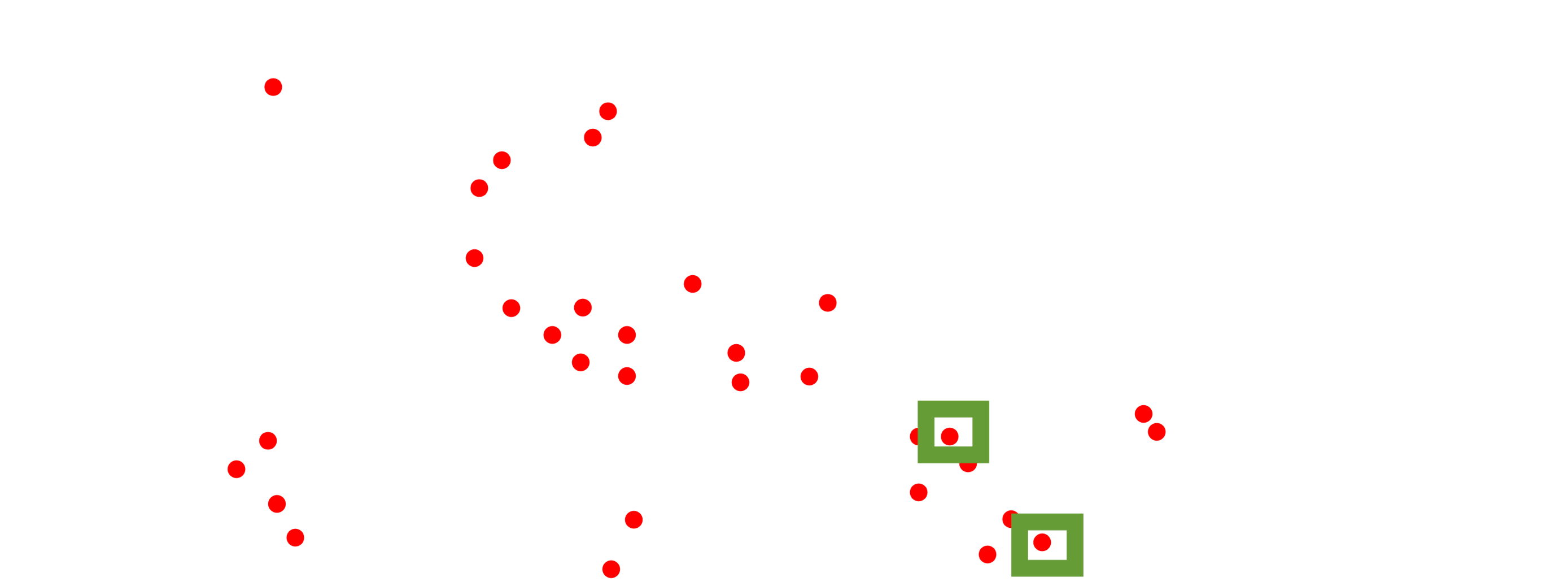}
       \caption{Our model.}
    \label{fig:ground_truth_2}
  \end{subfigure} 
  \begin{subfigure}{0.27\textwidth}
    \centering
    \includegraphics[width=\textwidth]{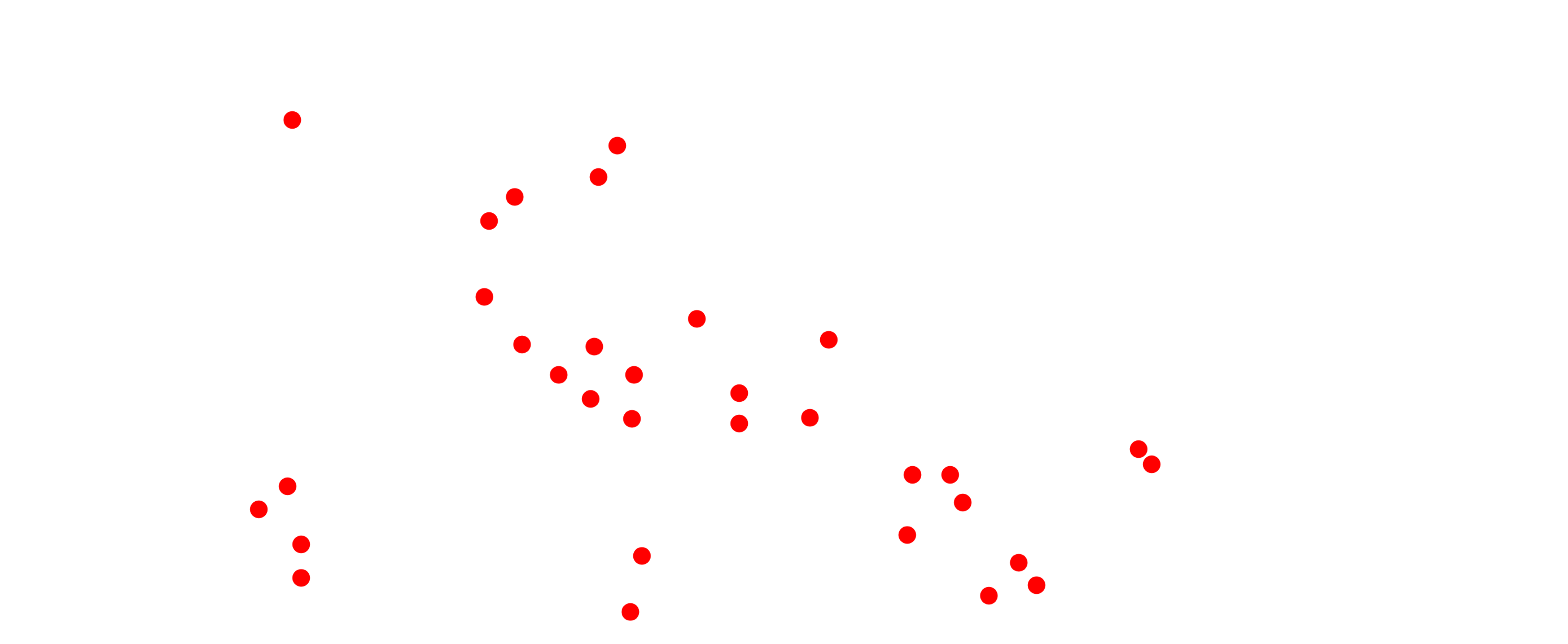}
   \caption{Ground truth.}
    \label{fig:third_image_2}
  \end{subfigure}
  \caption{Comparison of detection performance between our model and the baseline which does not incorporate the PVH on Wildtrack dataset, showing the predicted heatmap and the corresponding localization map after applying NMS and thresholding to the predicted heatmap.}
  \label{fig:detection_vis}
\end{figure*}
\subsection{Ablation Studies}

\textbf{Influence of different visual hull techniques.} We compare the performance of the standard visual hull (VH) and the Probabilistic Visual Hull (PVH) in Table~\ref{tab:ablation_study2} to compute the occupancy volume. Specifically, using the PVH increased MODA from $97.2\%$ to $97.3\%$, highlighting a slight improvement in pedestrian detection accuracy. Moreover, the PVH also led to a slight enhancement in both MODP, Precision, and Recall, with overall performance being close, but slightly better than, that of the traditional VH.

\medskip

\noindent \textbf{Influence of different PVH integration techniques.}  We evaluate the impact of various PVH integration methods in Table~\ref{tab:ablation_study1}. The first row presents the performance of the baseline model, which creates the 3D feature volume but does not incorporate the occupancy information. The second row shows the performance of the method where the PVH is concatenated with the 3D feature volume, leading to improved performance. The third row shows the performance of using the multiplication alone to integrate the PVH, without retaining the 3D feature volume. This results in the worst performance, with a significant drop in MODA. The fourth row illustrates the performance of adding the product of the PVH and the 3D feature volume back to the 3D feature volume, boosting the pedestrian features and yielding enhanced performance compared to the PVH concatenation alone. Notably, the best performance is achieved when the PVH is integrated through concatenation, shown in the last row, where the PVH is first multiplied with the 3D feature volume, and the result is then concatenated with the 3D feature volume, providing the most effective enhancement.

\subsection{Qualitative Analysis}

\begin{figure*}[t]
  \centering
  % Top row with three images
  \begin{subfigure}{0.25\textwidth}
    \centering
    \includegraphics[width=\textwidth]{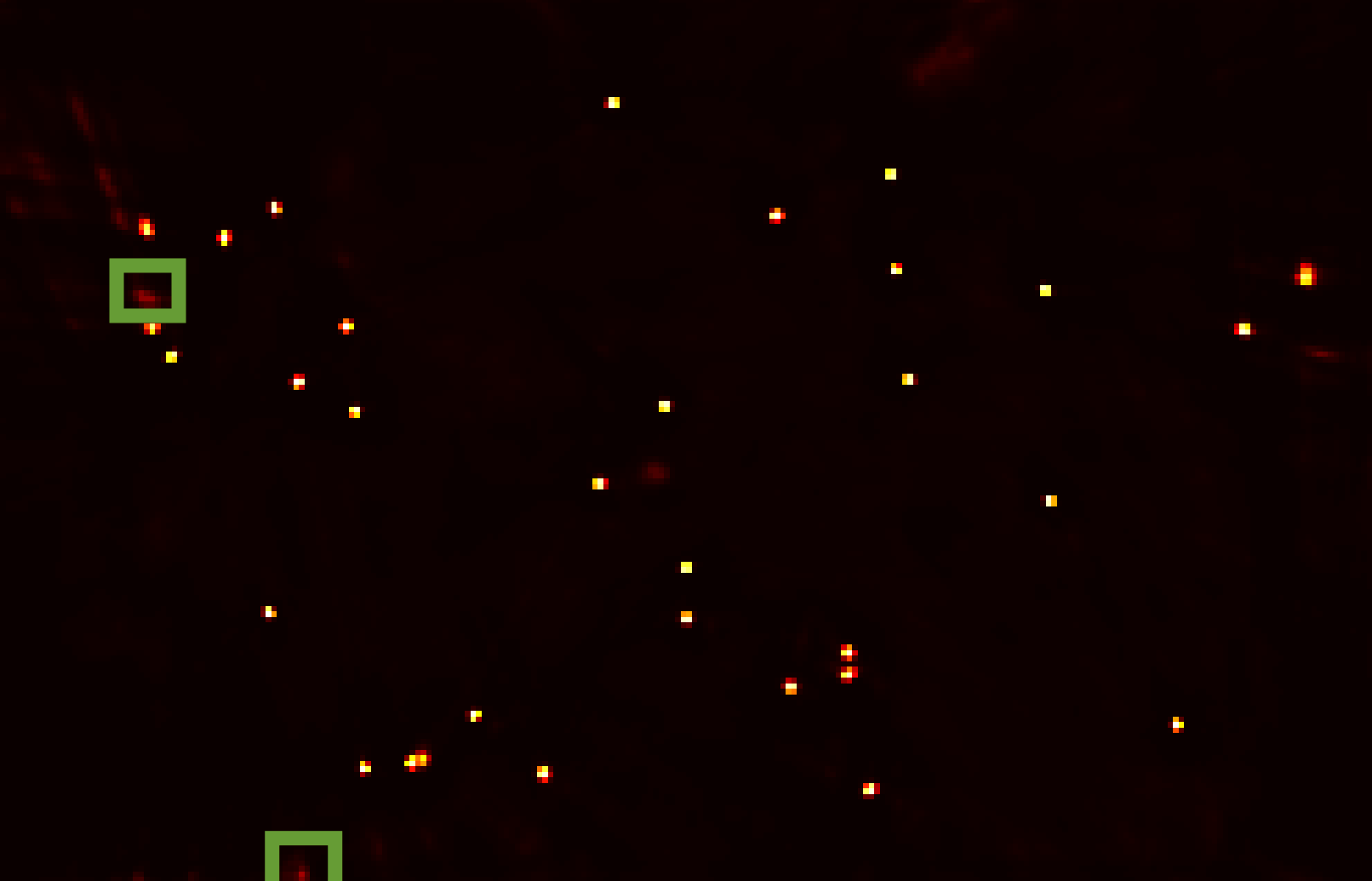}
    \label{fig:prediction}
  \end{subfigure} 
  \begin{subfigure}{0.25\textwidth}
    \centering
    \includegraphics[width=\textwidth]{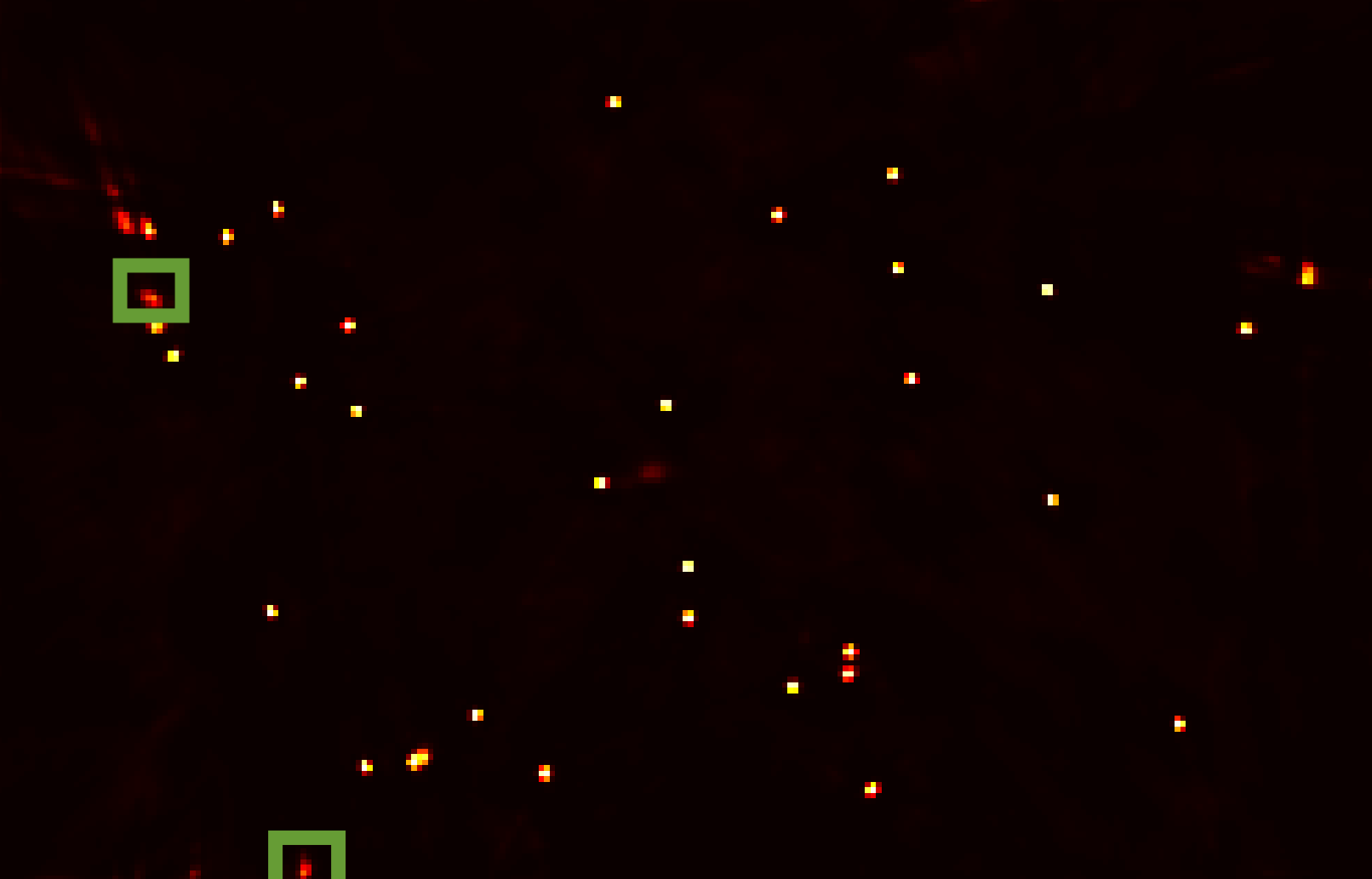}
    \label{fig:ground_truth}
  \end{subfigure} 
  \begin{subfigure}{0.25\textwidth}
    \centering
    \includegraphics[width=\textwidth]{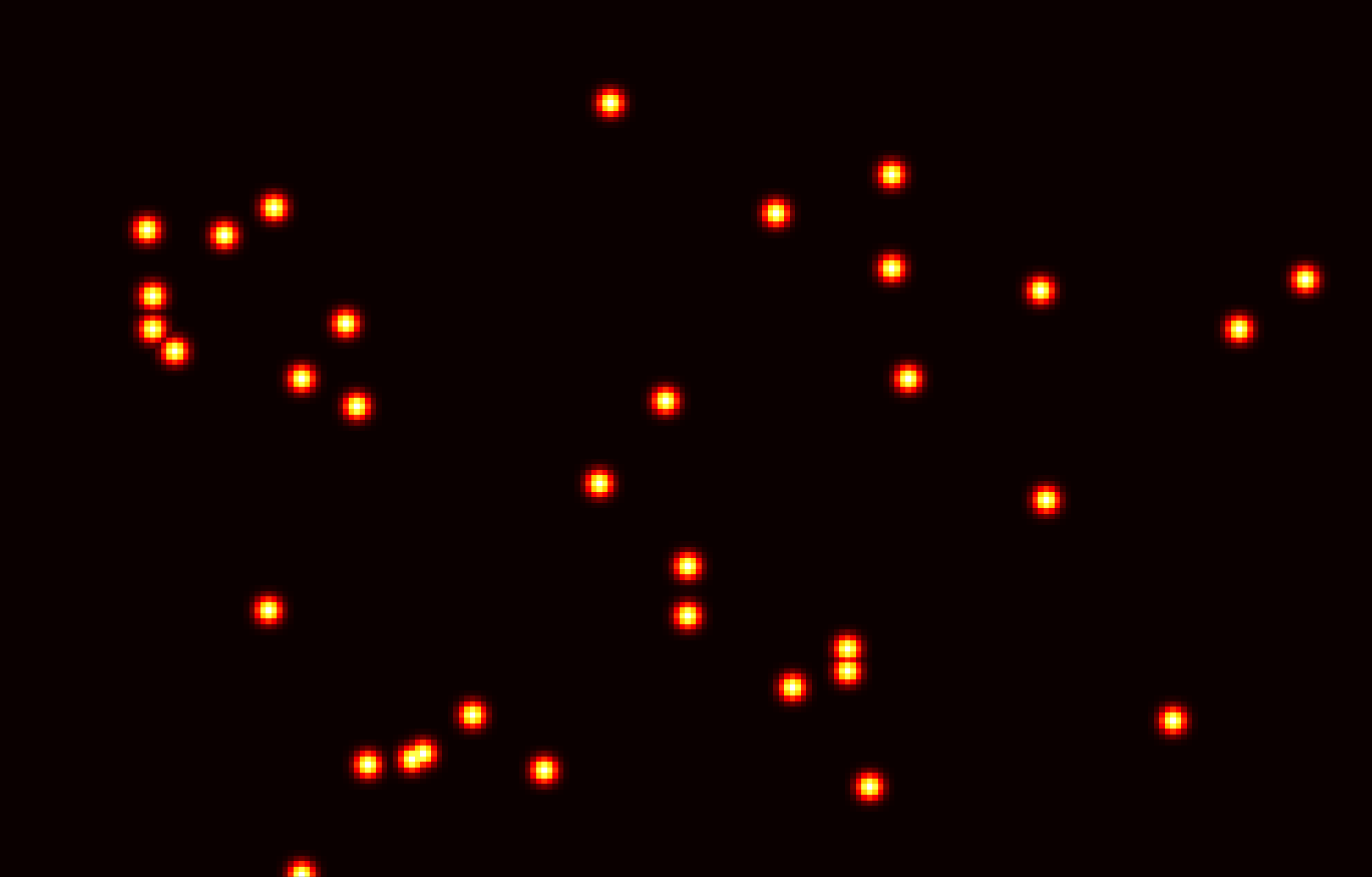}
    \label{fig:third_image}
  \end{subfigure}
  
  \begin{subfigure}{0.25\textwidth}
    \centering
    \includegraphics[width=\textwidth]{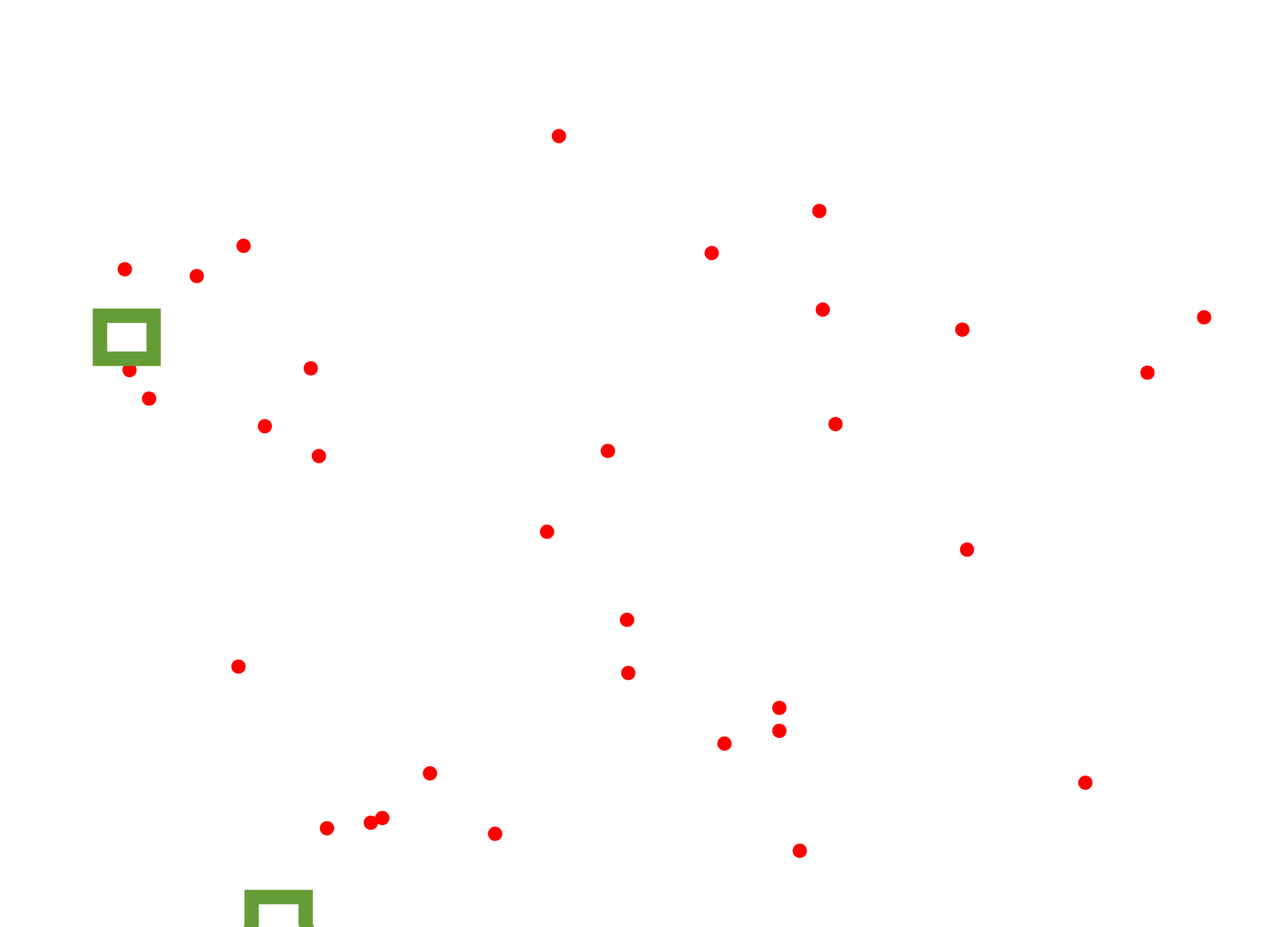}
    \label{fig:prediction_2}
  \end{subfigure} 
  \begin{subfigure}{0.25\textwidth}
    \centering
    \includegraphics[width=\textwidth]{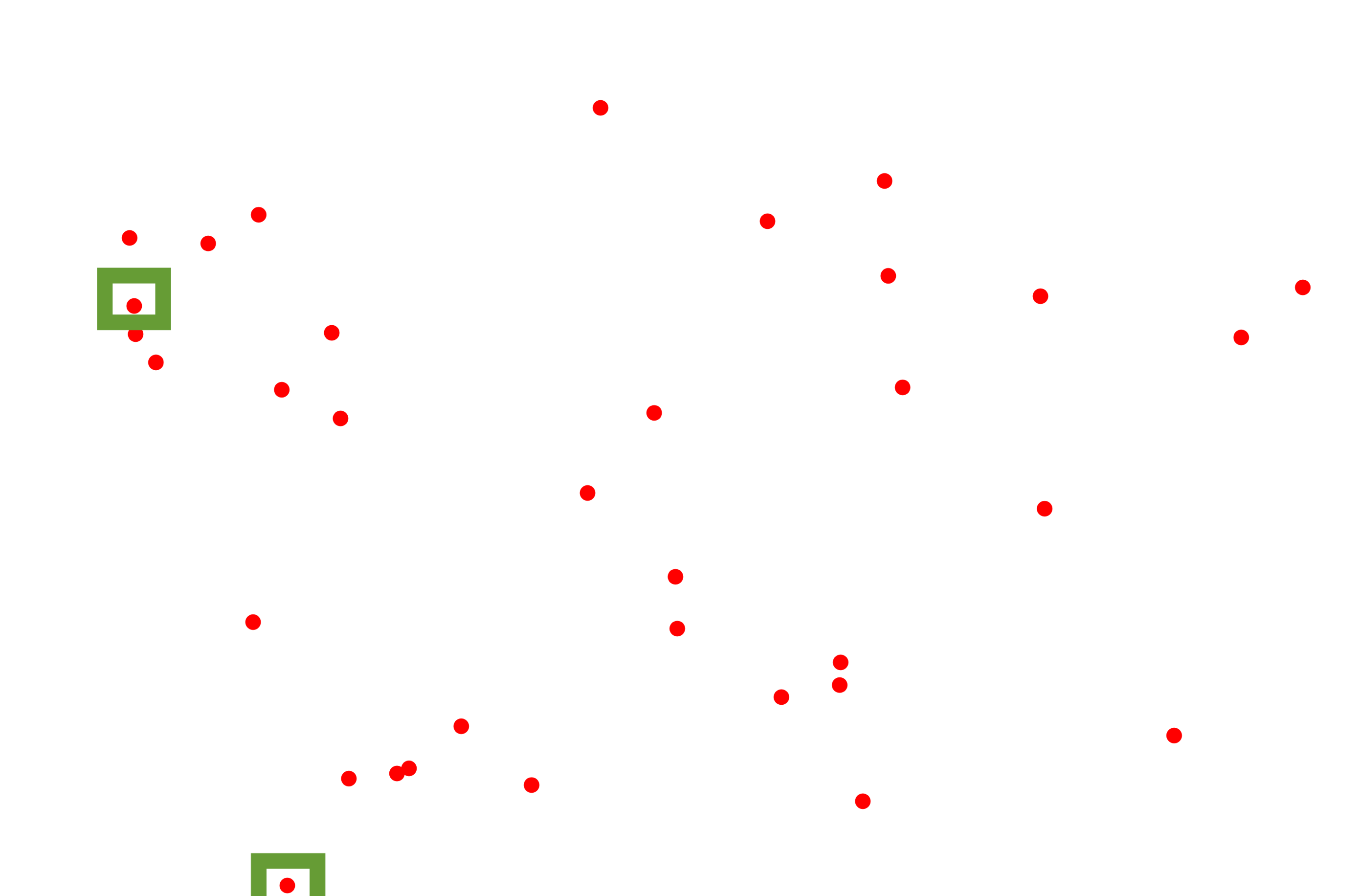}
    \label{fig:ground_truth_2}
  \end{subfigure} 
  \begin{subfigure}{0.25\textwidth}
    \centering
    \includegraphics[width=\textwidth]{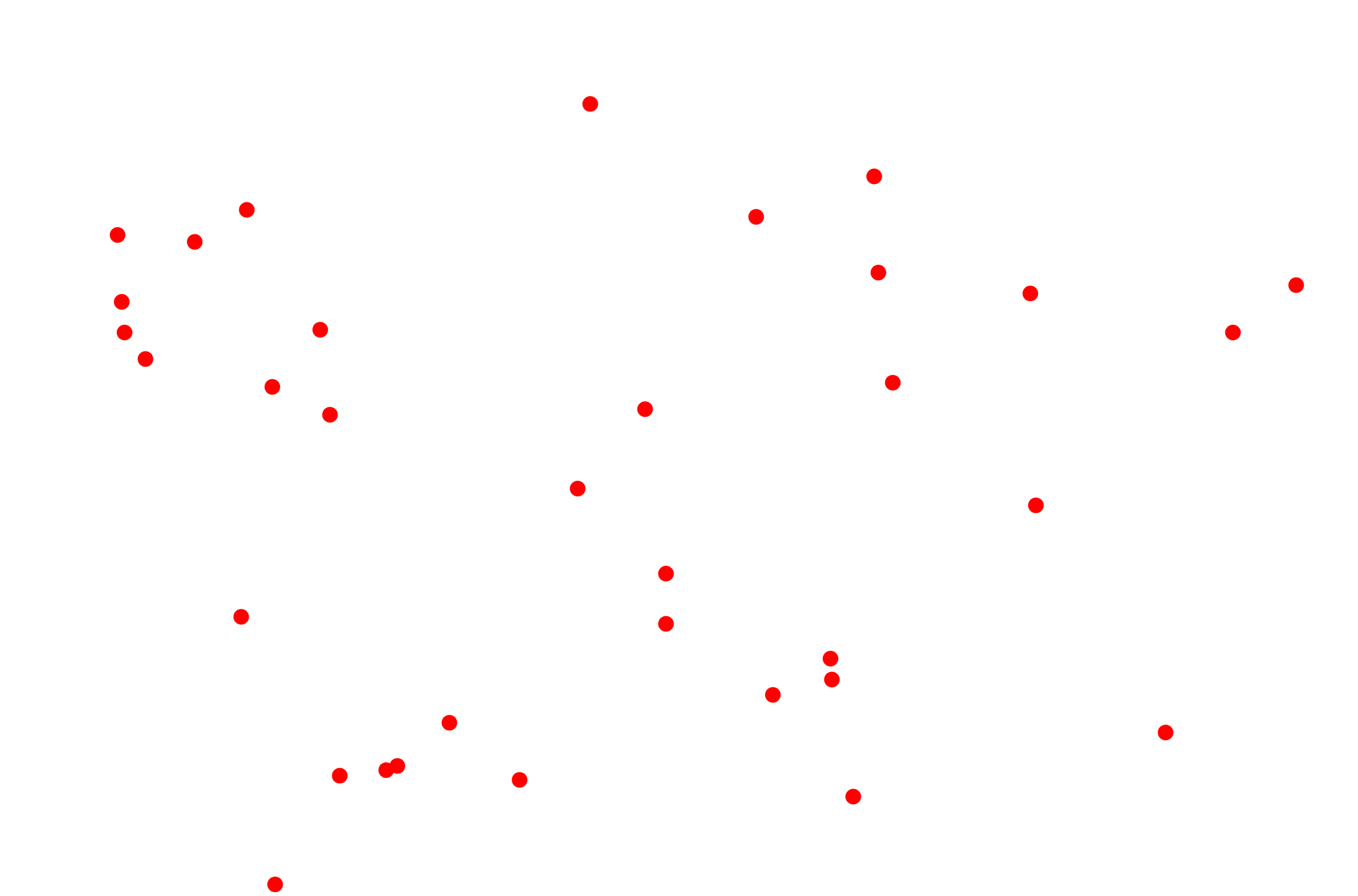}
    \label{fig:third_image_2}
  \end{subfigure}

  \vspace{-2mm} 
  \noindent\textcolor{gray}{\rule[0mm]{13.2cm}{0.1pt}} 
  \vspace{1mm}

    \centering
  % Top row with three images
  \begin{subfigure}{0.25\textwidth}
    \centering
    \includegraphics[width=\textwidth]{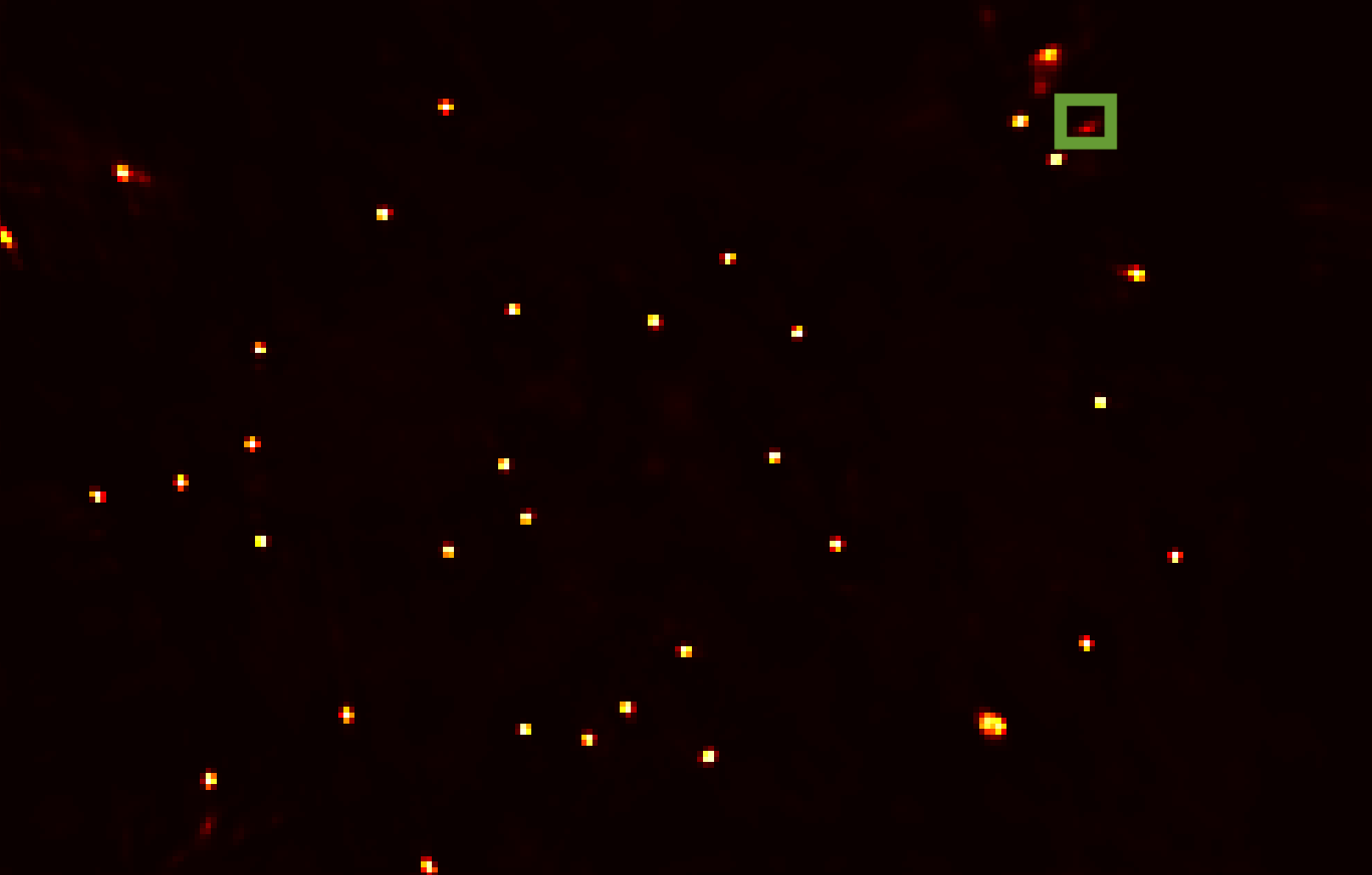}
    \label{fig:prediction}
  \end{subfigure} 
  \begin{subfigure}{0.25\textwidth}
    \centering
    \includegraphics[width=\textwidth]{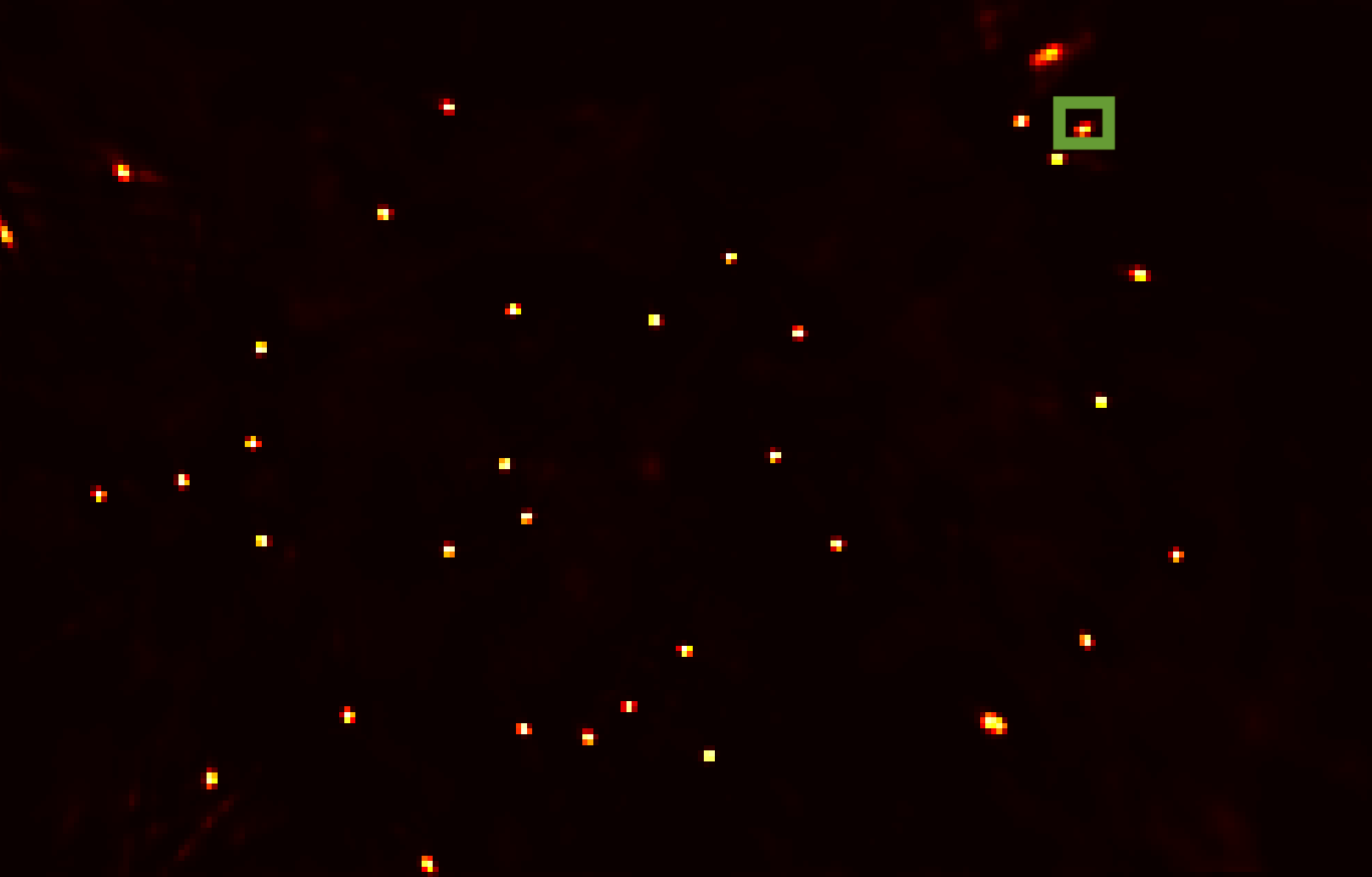}
    \label{fig:ground_truth}
  \end{subfigure} 
  \begin{subfigure}{0.25\textwidth}
    \centering
    \includegraphics[width=\textwidth]{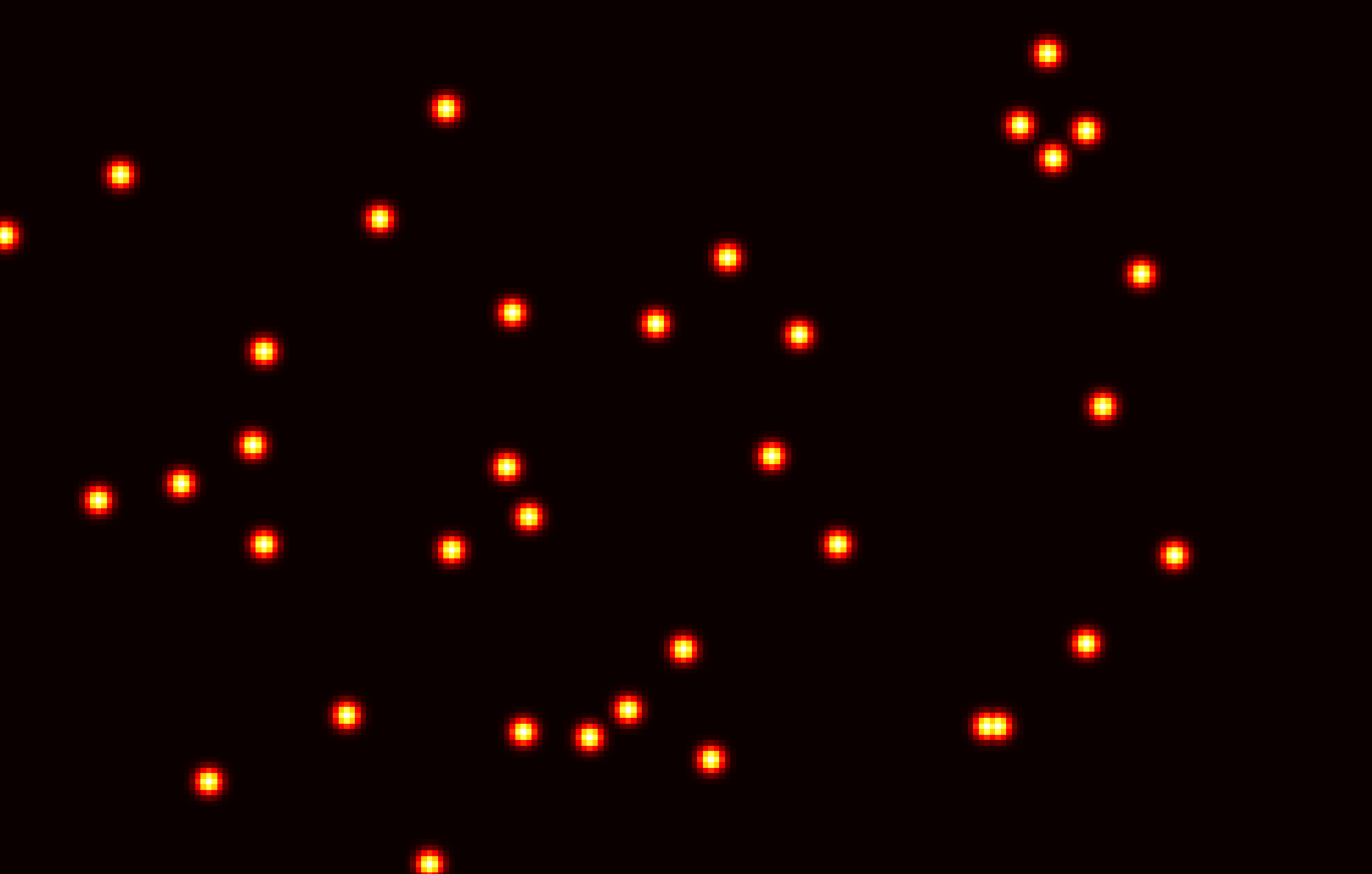}
    \label{fig:third_image}
  \end{subfigure}

  \begin{subfigure}{0.25\textwidth}
    \centering
    \includegraphics[width=\textwidth]{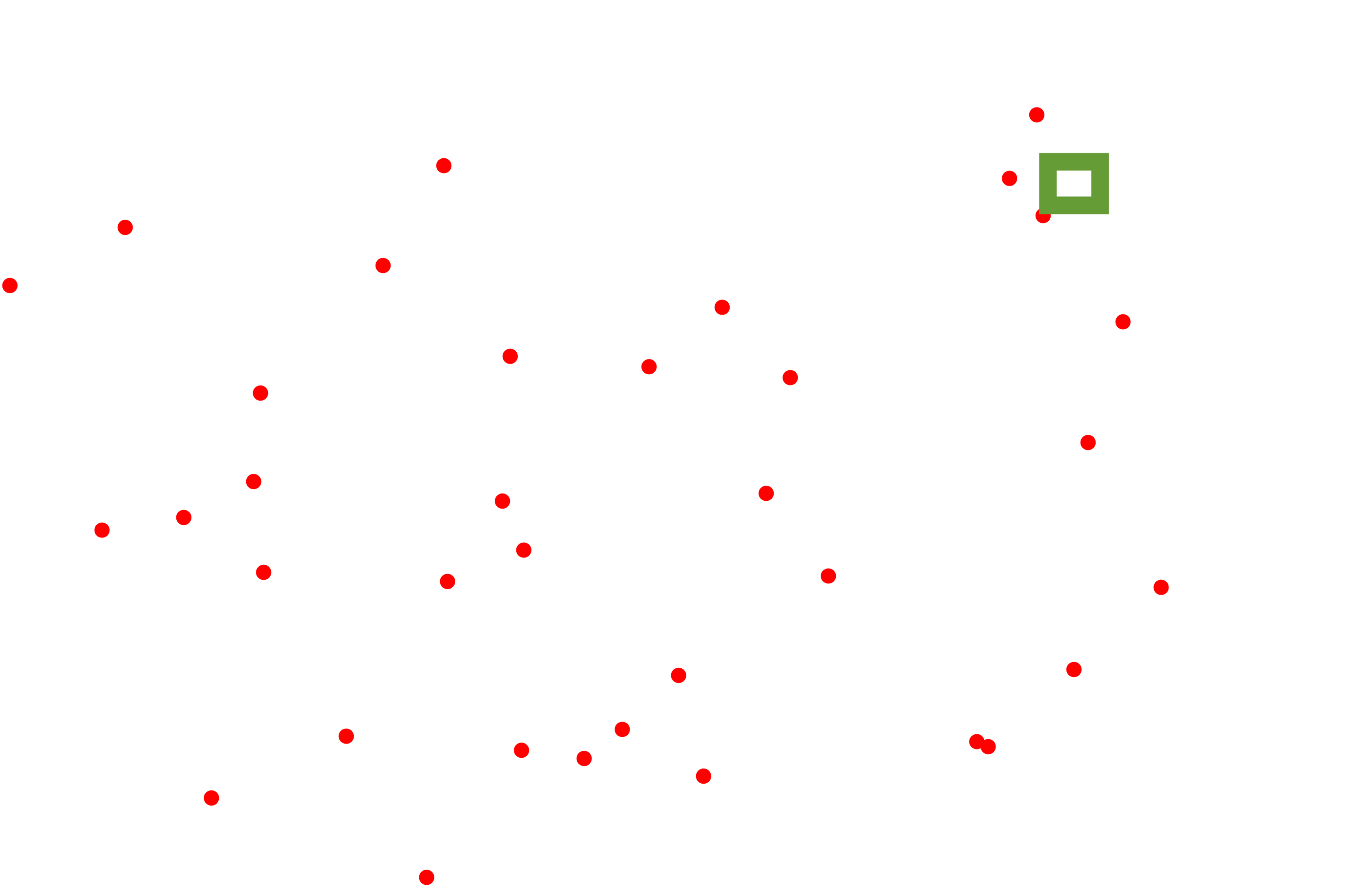}
    \caption{Baseline.}
    \label{fig:prediction_2}
  \end{subfigure} 
  \begin{subfigure}{0.25\textwidth}
    \centering
    \includegraphics[width=\textwidth]{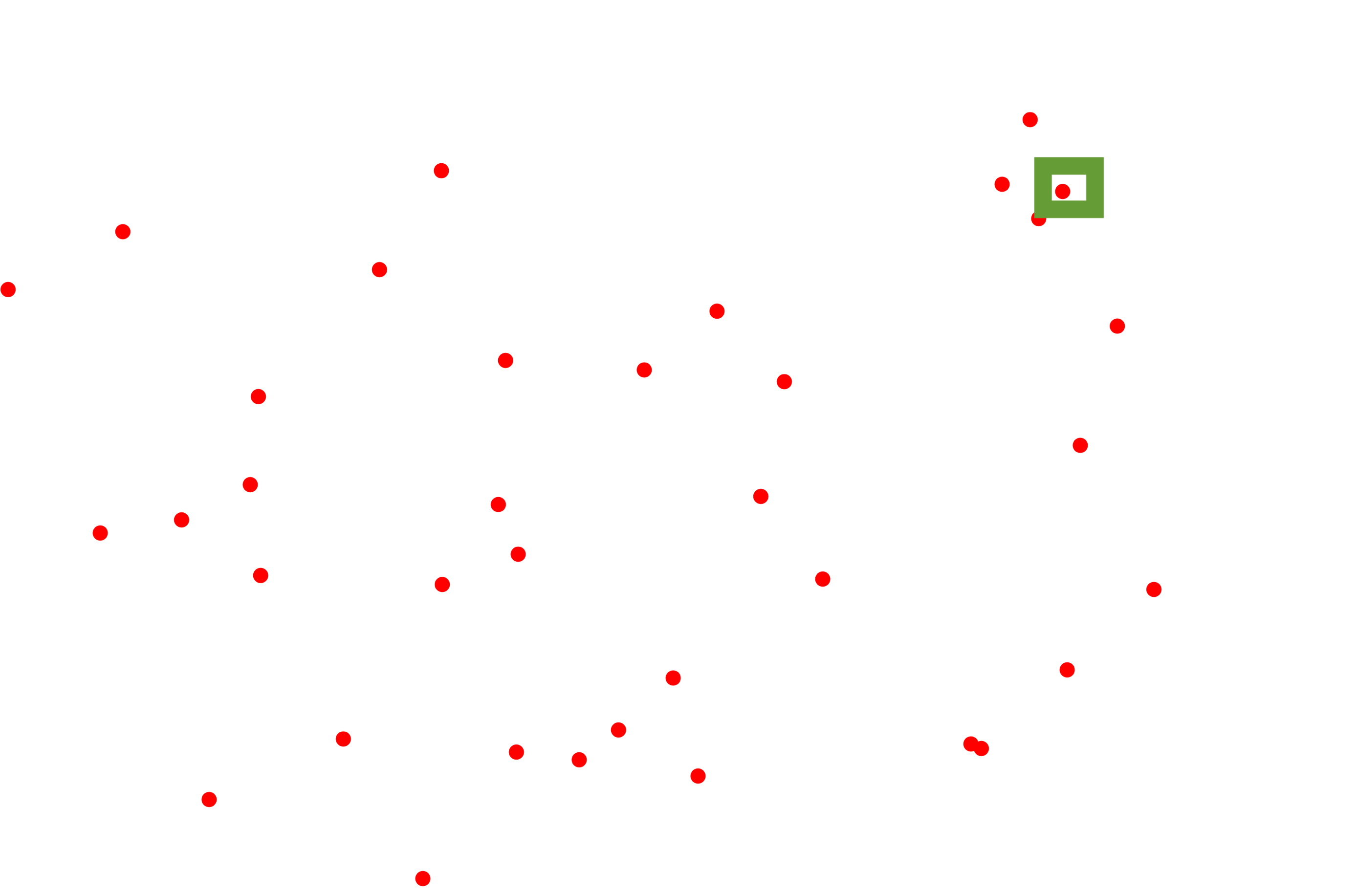}
       \caption{Our model.}
    \label{fig:ground_truth_2}
  \end{subfigure} 
  \begin{subfigure}{0.25\textwidth}
    \centering
    \includegraphics[width=\textwidth]{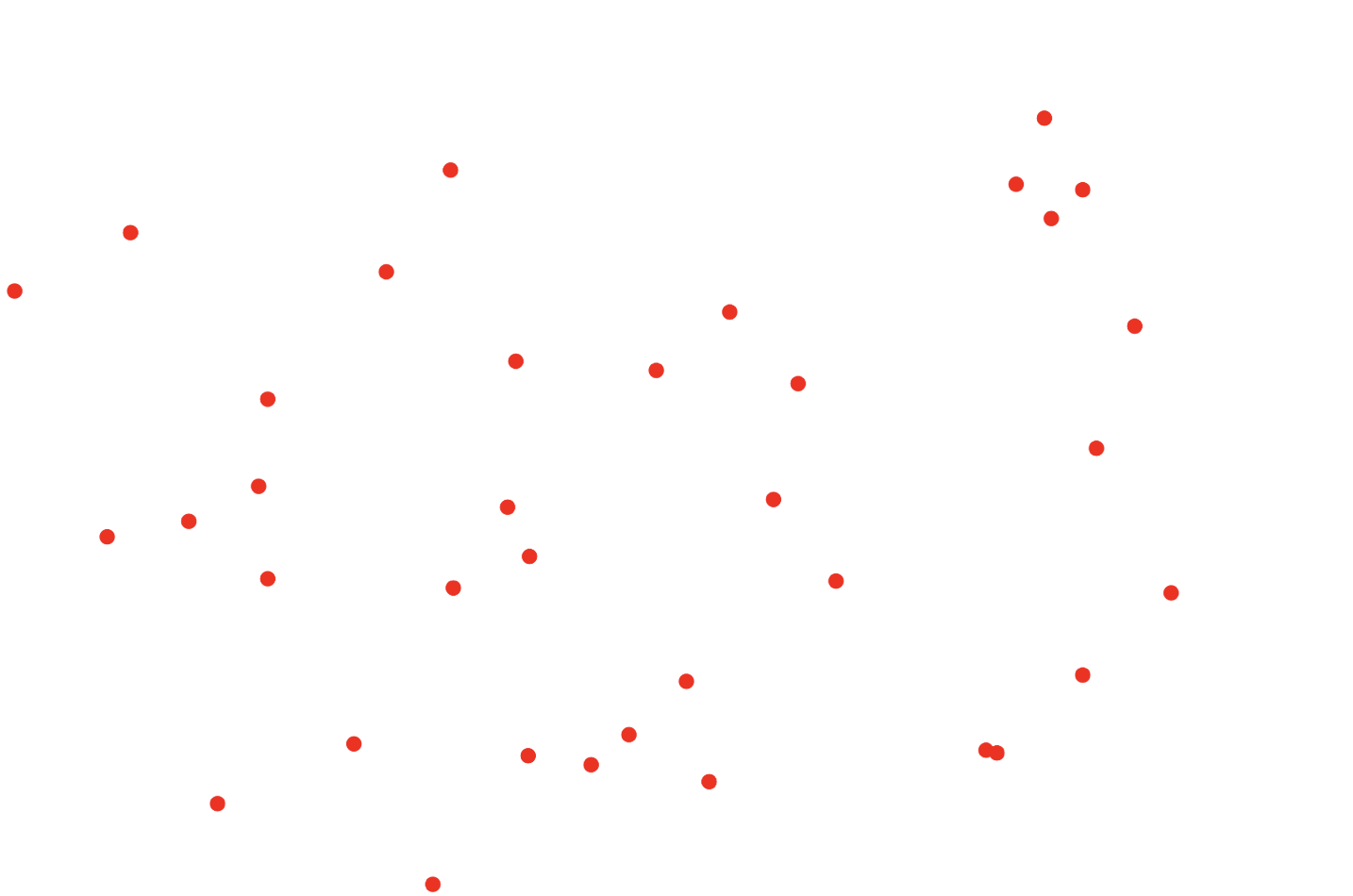}
   \caption{Ground truth.}
    \label{fig:third_image_2}
  \end{subfigure}

  \caption{Comparison of detection performance between our model and the baseline which does not incorporate the PVH on MultiviewX dataset, showing the predicted heatmap and the corresponding localization map after applying NMS and thresholding to the predicted heatmap.}
  \label{fig:detection_vis1}
\end{figure*}

To highlight the contribution of our proposed method, a qualitative comparison between the proposed model and the baseline is shown in Figures \ref{fig:detection_vis} and \ref{fig:detection_vis1}. Figure \ref{fig:detection_vis} presents the results using two frames from Wildtrack dataset, while Figure \ref{fig:detection_vis1} represnt the results using two frames from MultiviewX dataset. The predicted heatmap and corresponding localization map are visualized on the ground plane. We compare the results of our model, which incorporates the PVH, with the baseline method that uses the 3D feature volume and does not incorporate the PVH. The regions highlighted in green boxes illustrate the impact of incorporating the PVH, clearly showing how it enhances both the accuracy of the heatmap and the localization map, thus leading to better detection of pedestrians. Furthermore, it is worth noting that our model produces a more concentrated and less distorted detection heatmap on Wildtrack dataset, further emphasizing its superior performance.

\section{Conclusion}

We introduced a novel framework for enhancing multi-view pedestrian detection using simple 3D reconstruction techniques. We leverage the visual hull technique to provide a precise estimate of voxel occupancy by pedestrians, thereby enhancing detection accuracy while maintaining computational efficiency. Comprehensive experiments on the two benchmark datasets demonstrate the effectiveness of our proposed model. Additionally, through detailed analysis, we have identified key areas for future work that could further improve detection accuracy.

\section{Future Work}

While our model shows promising results, several areas remain for further improvement. One key challenge is addressing the issue of phantom volumes during visual hull reconstruction. These phantom volumes represent the reconstructed 3D regions that do not correspond to pedestrians. To address this issue, a potential direction is to explore alternative reconstruction methods that are not limited to silhouette information. On the other hand, although Mask R-CNN performs well in detecting pedestrians, it occasionally misidentifies non-pedestrians as pedestrians, affecting overall accuracy. Addressing these challenges will further refine the overall model's performance. Moreover, an interesting avenue for future work is to investigate the use of the proposed approach in a tracking context, exploring how volumetric occupancy can improve tracking accuracy.